\documentclass{article} 
\usepackage{iclr2027_conference,times}

\usepackage{amsmath,amsfonts,bm}

\def\eqref#1{equation~\ref{#1}}

\def\1{\bm{1}}

\DeclareMathAlphabet{\mathsfit}{\encodingdefault}{\sfdefault}{m}{sl}
\SetMathAlphabet{\mathsfit}{bold}{\encodingdefault}{\sfdefault}{bx}{n}

\usepackage{hyperref}
\usepackage{url}

\usepackage{graphicx}
\usepackage{booktabs}     
\usepackage{multirow}     
\usepackage{array}         
\usepackage{tabularx}      
\usepackage{makecell}     
\usepackage{amsmath}       
\usepackage{amssymb}
\usepackage{caption}
\usepackage{algorithm}
\usepackage{algpseudocode}
\usepackage{xcolor}
\usepackage{graphicx}
\usepackage{booktabs}
\usepackage{wrapfig}

\title{SERA: Scale-Equalized Rollout Allocation for Maximum Likelihood Reinforcement Learning}

\newcommand{\SERAauthorblock}{%
\parbox[t]{\dimexpr\textwidth-2\tabcolsep\relax}{%
\raggedright
\vspace*{0.4em}

{\bfseries
Zihao Chen\textsuperscript{1,\dag}
\hspace{0.7em}
Fanxiang Xiong\textsuperscript{2,\dag}
\hspace{0.7em}
Hongran Ren\textsuperscript{1}
\hspace{0.7em}
Xuefeng Bai\textsuperscript{1}
\hspace{0.7em}
Zhongxiang Dai\textsuperscript{2}
\\[0.2em]
Kehai Chen\textsuperscript{1}
\hspace{0.7em}
Zhiguo Zhang\textsuperscript{1}
\hspace{0.7em}
Zhiyong Wang\textsuperscript{1,*}
\hspace{0.7em}
Yu Cheng\textsuperscript{3}
\par}

\vspace{0.8em}

{\normalfont
\textsuperscript{1}Harbin Institute of Technology, Shenzhen
\\[0.2em]
\textsuperscript{2}The Chinese University of Hong Kong, Shenzhen
\\[0.2em]
\textsuperscript{3}Nanyang Technological University
\par}

}%
}

\author{\SERAauthorblock}

\newcommand{\method}{\textsc{SERA}}

\iclrfinalcopy 
\begin{document}

\maketitle
\pagestyle{plain}
\thispagestyle{plain}

\begin{NoHyper}
\renewcommand{\thefootnote}{}
\footnotetext{%
\textsuperscript{\dag}Equal contribution.
\quad
\textsuperscript{*}Corresponding authors.
}
\end{NoHyper}

\begin{abstract}
Maximum Likelihood Reinforcement Learning (MaxRL) targets prompt-wise log-success and has shown strong performance on reasoning tasks.
Under finite rollout budgets, however, the estimator used by MaxRL attenuates each prompt's likelihood gradient by a factor that depends on its success probability and rollout count.
Under uniform rollout allocation, the common rollout count fails to compensate for success-dependent attenuation, leaving low-success prompts more strongly attenuated and distorting their relative contributions to the expected aggregate gradient.
We introduce SERA (Scale-Equalized Rollout Allocation), which redistributes a fixed rollout budget to approximately equalize these finite-rollout scaling factors.
Building on our theoretical analysis of how finite rollouts distort prompt-wise likelihood gradients, we formulate the allocation as a fixed-budget max--min problem, derive a waterline solution to its continuous relaxation, and introduce a multiplicity correction to remove the additional prompt weighting induced by heterogeneous rollout counts.
Experiments show stronger alignment with exact likelihood gradients in a controlled ImageNet setting and improved multi-sample solution coverage over MaxRL on maze navigation and mathematical reasoning under matched training rollout budgets. Code is available at \url{https://github.com/ChenZihao0121/SERA}.
\end{abstract}

\section{Introduction}
Reinforcement learning with verifiable rewards has become a standard approach to improving language-model reasoning, as exemplified by GRPO~\citep{shao2024deepseekmath} and DeepSeek-R1~\citep{guo2025deepseekr1}. Yet single-sample accuracy does not fully characterize reasoning performance: at inference time, models are often sampled repeatedly, and their ability to produce at least one correct solution is commonly measured by Pass@$K$~\citep{chen2021evaluating}. Improvements in Pass@1 need not translate into better coverage at larger sampling budgets~\citep{yue2025reasoningcapacity}, motivating objectives that account for multi-sample success. Maximum Likelihood Reinforcement Learning (MaxRL)~\citep{tajwar2026maxrl} targets average prompt-wise log-success, placing greater relative weight on improvements for low-success prompts. Its likelihood gradient can be expressed as a combination of Pass@$K$ gradients across sampling budgets, directly connecting the objective to multi-sample solution coverage.

Realizing this objective with finitely many rollouts, however, introduces a systematic mismatch between the target likelihood gradient and the expected MaxRL update.
The practical centered MaxRL estimator attenuates each prompt's log-success gradient by a coefficient that depends jointly on its success probability and rollout count. Under uniform allocation, the same rollout count leaves low-success prompts more strongly attenuated, reducing their relative contributions despite the greater emphasis placed on them by the target log-success objective.
When prompt gradients differ in direction, such unequal scaling can also shift the expected batch gradient away from the target
likelihood gradient.
This raises a central question: \emph{Given a fixed training rollout budget, how should rollouts be
distributed across prompts to better preserve their relative
log-success gradient contributions?}

Existing rollout allocation methods adapt sampling based on prompt difficulty, learning signals, or objective-induced weighting~\citep{yang2025dars,xiong2025reinforceada,nguyen2026vip,li2025knapsack}. In contrast, we study the prompt-dependent finite-rollout scaling induced by the centered MaxRL estimator itself. Specifically, we ask how a fixed rollout budget should be allocated to control the coefficients that scale prompt-wise likelihood gradients in the expected MaxRL update.

We introduce \method{} (Scale-Equalized Rollout Allocation), which redistributes a fixed rollout budget to approximately equalize the finite-rollout scaling coefficients across prompts. For a fixed retained prompt set, our analysis shows that matching these coefficients preserves the relative likelihood-gradient contributions up to a common factor. We therefore maximize the minimum scaling coefficient across prompts under the budget constraint; in the continuous relaxation, the optimum equalizes these coefficients and admits a waterline solution. The resulting allocation recovers inverse-success behavior in the low-success regime while retaining the full finite-rollout dependence of the centered MaxRL estimator. Unequal rollout counts, however, also induce additional prompt weighting under response-mean aggregation. We therefore pair the allocation with a multiplicity correction that removes this effect, allowing rollout counts to control finite-rollout attenuation without introducing a separate group-size weighting.

We evaluate \method{} against MaxRL under matched training rollout budgets on ImageNet~\citep{deng2009imagenet}, maze navigation~\citep{tajwar2026maxrl}, and mathematical reasoning with SmolLM2-360M-Instruct~\citep{allal2025smollm2}, Qwen2.5-Math-1.5B~\citep{yang2024qwen25math}, and Qwen3-4B-Base~\citep{yang2025qwen3}. In a controlled ImageNet setting, \method{} produces sampled updates that align more closely with the exact likelihood gradient, supporting the predicted effect of scale equalization. Using historical success estimates, \method{} also improves held-out maze coverage under both tested loss normalizations and consistently increases multi-sample solution coverage on mathematical reasoning tasks. On SmolLM2-360M-Instruct, Pass@256 on GSM8K-Platinum~\citep{cobbe2021gsm8k,vendrow2025reliability} improves by $5.13$ percentage points under both normalization schemes. On Qwen2.5-Math-1.5B, Pass@512 improves by $7.00$ points on BeyondAIME~\citep{bytedanceseed2025beyondaime} and $6.67$ points on AIME 2025~\citep{zhang2025aime25}; on Qwen3-4B, it improves by $8.00$ points on BeyondAIME and $4.78$ points on Minerva Math~\citep{lewkowycz2022minerva}.

Our contributions are:
\begin{enumerate}
    \item \textbf{Finite-rollout analysis of the centered MaxRL estimator.}
    We characterize the conditional objective and expected batch
    gradient of centered MaxRL under heterogeneous rollout counts,
    showing how unequal prompt-wise scaling can alter the aggregate
gradient direction and how scale equalization preserves
    the relative likelihood-gradient contributions of a fixed
    retained set.

    \item \textbf{Scale-equalized rollout allocation.}
    Building on this analysis, we develop \method{}, formulating
rollout allocation as a fixed-budget max--min problem over the
finite-rollout scaling coefficients and deriving a waterline
solution to its continuous relaxation.
We further introduce a multiplicity correction that removes
the additional prompt weighting induced by unequal rollout
counts under response-mean aggregation.

    \item \textbf{Experimental validation across tasks.}
    We validate the gradient-alignment effect predicted by our analysis in a controlled ImageNet setting and demonstrate improved multi-sample solution coverage over MaxRL on maze navigation and mathematical reasoning with language models under matched training rollout budgets.
\end{enumerate}

\section{Preliminaries}
\label{sec:Prel}

We consider generation tasks with binary verifier rewards
$R_q(y)\in\{0,1\}$.
For prompt $q$ and response
$y\sim\pi_\theta(\cdot\mid q)$, define
$p_q(\theta):=\Pr[R_q(y)=1]$.
Following MaxRL~\citep{tajwar2026maxrl}, we consider the
prompt-wise log-success objective
\begin{equation}
    J_{\log}(\theta)
    :=
    \mathbb E_{q\sim\mathcal D}[\log p_q(\theta)].
    \label{eq:log-success}
\end{equation}

Given $N$ i.i.d.\ responses
$y_{qi}\sim\pi_\theta(\cdot\mid q)$, let
$R_{qi}=R_q(y_{qi})$,
$\overline R_q=\frac{1}{N}\sum_{i=1}^{N}R_{qi}$, and
$s_{qi}=\nabla_\theta\log\pi_\theta(y_{qi}\mid q)$.
To reduce gradient variance and improve training stability,
practical MaxRL uses the following centered estimator:
\begin{equation}
    \widehat g_q^{(N)}
    :=
    \begin{cases}
        \displaystyle
        \frac1N\sum_{i=1}^{N}
        \frac{R_{qi}-\overline R_q}{\overline R_q}\,s_{qi},
        & \overline R_q>0,\\[8pt]
        0,
        & \overline R_q=0.
    \end{cases}
    \label{eq:practical-maxrl}
\end{equation}

For $p_q\in(0,1)$, its expectation is
\begin{equation}
    \mathbb E[\widehat g_q^{(N)}]
    =
    \underbrace{
        \left[1-(1-p_q)^{N-1}\right]
    }_{\kappa(p_q,N)}
    \nabla_\theta\log p_q.
    \label{eq:finite-rollout-fidelity}
\end{equation}
We refer to $\kappa(p,N)$ as the \emph{finite-rollout fidelity}. At a fixed rollout count, this coefficient increases with the success probability, so lower-success prompts experience stronger attenuation. Uniform rollout allocation can therefore produce unequal scaling
across prompts.
Since prompt gradients need not be collinear, this can
change the direction of their expected aggregate.
Further theoretical details are provided in
Appendix~\ref{app:additional-theory}.

\begin{figure}[t]
    \centering
    \includegraphics[width=\linewidth]
    {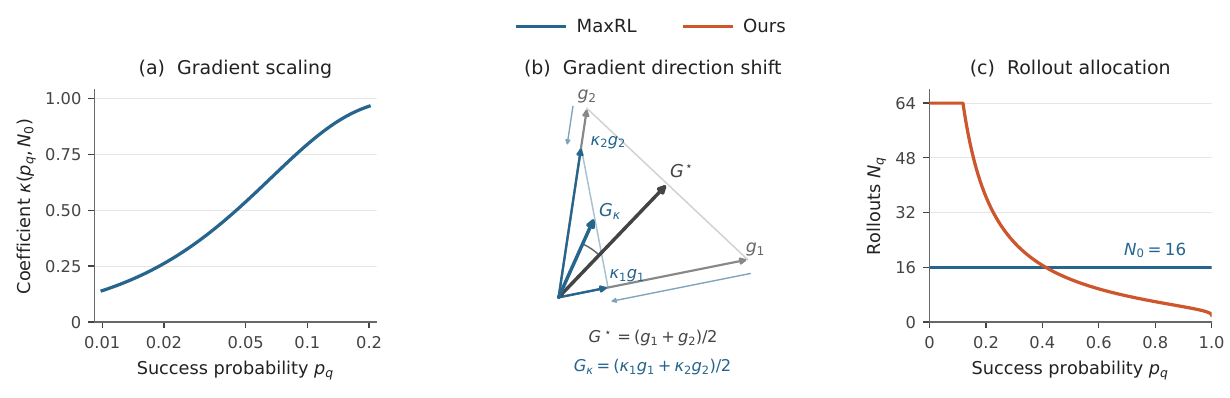}
    \caption{
    \textbf{Gradient scaling and rollout allocation.}
    (a) Fixed-rollout MaxRL induces success-dependent scaling. 
    (b) Unequal prompt-wise scaling can alter the expected
    batch-gradient direction.
    (c) Ours assigns more rollouts to lower-success prompts
    under the same total budget ($N_{\max}=64$).}
    \label{fig:allocation-principle}
\end{figure}

\section{Method}
\label{sec:method}

\method{} redistributes a fixed rollout budget to approximately equalize the prompt-wise finite-rollout scaling coefficients of the
centered MaxRL estimator.
The resulting allocation is paired with multiplicity correction
to remove the additional response-count weighting introduced
by unequal rollout counts under response-mean aggregation.

\subsection{Rollout Allocation}
\label{sec:allocation-motivation}

Equation~\ref{eq:finite-rollout-fidelity} shows that
the expected MaxRL estimator is scaled by
$\kappa(p_q,N)=1-(1-p_q)^{N-1}$.
As illustrated in Figure~\ref{fig:allocation-principle} (a),
under a fixed rollout count $N_0$, this coefficient
varies with the success probability $p_q$.
These unequal coefficients change the relative
contributions of prompt gradients and can therefore
alter the direction of their expected aggregate.
Figure~\ref{fig:allocation-principle} (b) illustrates this
effect: although scaling preserves each prompt
gradient's direction, their expected mean deviates
from the uniformly averaged log-success gradient.

If the success probabilities were known, directly dividing
each estimator by $\kappa(p_q,N_0)$ would remove
this scaling in expectation.
However, this correction leaves the probability of
zero-signal groups unchanged and multiplies the
estimator's covariance by $1/\kappa(p_q,N_0)^2$,
amplifying noise when $\kappa$ is small.
We instead adjust rollout counts to equalize
$\kappa(p_q,N_q)$ across prompts.
A shared positive coefficient preserves the relative log-success gradient contributions up to a common factor, while changing the rollout counts also changes the probability of obtaining mixed-reward groups.

Let $\mathcal B$ contain $B$ prompts, with a fixed
total rollout budget $M:=BN_0$ and $N_0\ge2$.
Under this fixed budget, we maximize the minimum fidelity coefficient across prompts, thereby improving the most strongly attenuated prompt.
Initially treating counts as continuous and setting
aside practical count bounds, we consider
\begin{align}
    \max_{\{N_q\}_{q\in\mathcal B}}
    \quad &\min_{q\in\mathcal B}\kappa(p_q,N_q)
    \label{eq:waterline-objective}\\
    \text{s.t.}\quad
    &\sum_{q\in\mathcal B}N_q=M,
    \qquad N_q\ge 1.
    \nonumber
\end{align}
Because $\kappa(p_q,N_q)$ increases with $N_q$, any continuous allocation with unequal fidelity levels can increase the minimum by shifting rollout budget from a higher-fidelity prompt to a lower-fidelity one. The continuous optimum therefore equalizes the coefficients across prompts.
Writing their common value as $\kappa_{\mathcal B}^\star$
and defining
$c_{\mathcal B}:=-\log(1-\kappa_{\mathcal B}^\star)$,
we obtain
$(N_q^\star-1)[-\log(1-p_q)]=c_{\mathcal B}$.
The resulting allocation is
\begin{equation}
    N_q^\star
    =
    1+\frac{c_{\mathcal B}}{-\log(1-p_q)}.
    \label{eq:closed-form-allocation}
\end{equation}
The budget constraint then gives
\begin{equation}
    c_{\mathcal B}
    =
    \frac{M-B}
    {\displaystyle
     \sum_{q\in\mathcal B}[-\log(1-p_q)]^{-1}},
    \qquad
    \kappa_{\mathcal B}^\star=1-e^{-c_{\mathcal B}}.
    \label{eq:shared-allocation-constant}
\end{equation}
Thus, lower-success prompts receive more rollouts
to compensate for stronger attenuation.
The shared water level $c_{\mathcal B}$ depends
on the entire batch, and equal success probabilities
recover uniform allocation. In practice, rollout counts must be integers satisfying
$N_{\min}\le N_q\le N_{\max}$.
We first incorporate these bounds into the continuous
allocation:
\begin{equation}
    N_q^{\mathrm{cont}}
    =
    \operatorname{clip}\!\left(
        1+\frac{c_{\mathcal B}}{-\log(1-p_q)},
        N_{\min},N_{\max}
    \right),
    \qquad
    \sum_{q\in\mathcal B}N_q^{\mathrm{cont}}=M.
    \label{eq:centered-waterline}
\end{equation}
Here, $c_{\mathcal B}$ is a shared constant recomputed
so that the bounded counts sum to $M$.
This waterline solves the bounded continuous max--min
problem (Appendix~\ref{app:waterline}).
Figure~\ref{fig:allocation-principle} (c) illustrates
this allocation: under the same total budget as
fixed-rollout MaxRL, lower-success prompts receive
more rollouts, up to $N_{\max}$.
We then round the counts to integers while preserving
the budget and bounds, yielding approximate scale
equalization. The integerization procedure is detailed
in Appendix~\ref{app:integerization}.

\subsection{Observability Safeguard}
\label{sec:active-curriculum}

Equalizing gradient scales does not ensure that a
sampled group provides a nonzero learning signal.
For the centered estimator, both all-incorrect and
all-correct groups have zero advantages.
Nonzero advantages therefore require a group to
contain both a success and a failure, which occurs
with probability
\begin{equation}
    U_q(N):=1-(1-p_q)^N-p_q^N.
    \label{eq:mixed-group-probability}
\end{equation}
Using estimates $\widehat p_q$, we rank
prompts by $\widehat U_q(N_{\max})$ and search over
capacity-feasible prefixes of this ranking.
For each candidate set, we recompute the waterline
under the original budget $M=BN_0$, obtain integer
counts using Section~\ref{sec:allocation-motivation},
and check whether every retained prompt satisfies
$\widehat U_q(N_q)\ge u_0$, where $u_0\in[0,1)$.
If no tested prefix qualifies, a capacity-preserving
fallback allocates the full budget to the top-ranked
prompts (Appendix~\ref{app:active-set-details}).
The selected set $\mathcal A\subseteq\mathcal B$
receives the full original budget, while inactive
prompts receive zero rollouts.

\begin{algorithm}[t]
\caption{Training procedure for \method{}}
\label{alg:method}
\begin{algorithmic}[1]
\Require Policy $\pi_\theta$; baseline count $N_0$;
         bounds $N_{\min},N_{\max}$; threshold $u_0$
\Require Prior $\alpha_0,\beta_0>0$;
         discount $\lambda\in[0,1]$
\State Initialize historical statistics and pending outcomes
\For{each epoch}
    \State Update discounted statistics and freeze
           available success estimates
    \State Clear pending outcomes
    \For{candidate batch $\mathcal B$}
        \State $M\gets|\mathcal B|N_0$
        \If{all prompts have available estimates}
            \State Jointly select $\mathcal A$ and integer
                   counts using observability-guided selection
                   and bounded waterline Equation~\ref{eq:centered-waterline}
        \Else
            \State $\mathcal A\gets\mathcal B$;
                   $N_q\gets N_0$ for all $q\in\mathcal B$
        \EndIf
        \State Generate $N_q$ responses for each
               $q\in\mathcal A$ and evaluate rewards
        \State Compute centered MaxRL advantages
               as in Equation~\ref{eq:practical-maxrl}
        \State Apply response weights from Equation~\ref{eq:prompt-weight}
               to the advantages and update $\theta$
        \State Accumulate observed rewards in pending outcomes
    \EndFor
\EndFor
\end{algorithmic}
\end{algorithm}

\subsection{Implementation}
\label{sec:practical-algorithm}

\paragraph{Success-probability estimation.}
Inspired by other allocation methods~\citep{li2025knapsack,xiong2025reinforceada},
we estimate $p_q$ using reward observations
from previous epochs.
At the start of epoch $e$, we compute
\begin{equation}
    \widehat p_q^{(e)}
    =
    \frac{\alpha_0+S_q^{(e)}}
    {\alpha_0+\beta_0+S_q^{(e)}+F_q^{(e)}},
    \label{eq:discounted-beta-estimate}
\end{equation}
where $S_q^{(e)}$ and $F_q^{(e)}$ are discounted
success and failure counts from earlier epochs,
and $\alpha_0,\beta_0>0$ define a fixed Beta prior.
Estimates remain fixed within each epoch, and new
outcomes are incorporated at the next epoch boundary.

\paragraph{Multiplicity correction.}
Under response-mean aggregation, directly averaging
the losses of all generated responses weights each
prompt in proportion to its rollout count $N_q$.
Even when the coefficients $\kappa(p_q,N_q)$ are
equalized, this additional weighting can change
the relative contributions of prompt gradients.
To remove this response-count weighting, we multiply
each response's centered advantage by
\begin{equation}
    w_q:=\frac{N_0}{N_q},
    \qquad q\in\mathcal A.
    \label{eq:prompt-weight}
\end{equation}
Since $N_qw_q=N_0$, every retained prompt has the
same total response weight, regardless of its
rollout count.
Under response-mean aggregation, this preserves
the uniform weighting of retained prompts used
in the allocation analysis.
\paragraph{Training procedure.}
Algorithm~\ref{alg:method} summarizes the training
procedure. For each candidate batch, we combine the rollout allocation in Section~\ref{sec:allocation-motivation} with the observability safeguard in Section~\ref{sec:active-curriculum} to select the retained set and determine its integer rollout counts under the original budget.
If any candidate prompt lacks an available estimate,
the batch uses uniform counts $N_0$.
We then generate the assigned responses, compute
centered advantages, and apply the response weights
before the policy update.
Further implementation details are provided in
Appendix~\ref{app:implementation-details}.

\section{Experiments}
\label{sec:experiments}

We evaluate \method{} through controlled gradient
comparisons, sequence generation, and mathematical
reasoning.
We begin with ImageNet~\citep{deng2009imagenet}
classification (Section~\ref{sec:exp-imagenet}),
where exact maximum-likelihood gradients provide a controlled reference for assessing finite-rollout updates.
Using exact success probabilities for allocation, we isolate the effect of rollout allocation from success-probability estimation error and examine both gradient alignment and held-out performance.
We then study maze navigation with a fixed training
set (Section~\ref{sec:exp-maze}), extending the comparison
to sequence generation with binary verifier-based rewards.
Finally, we evaluate mathematical reasoning with
SmolLM2-360M-Instruct~\citep{allal2025smollm2}
(Section~\ref{sec:exp-smollm}).
We further evaluate
Qwen2.5-Math-1.5B~\citep{yang2024qwen25math}
and Qwen3-4B-Base~\citep{yang2025qwen3}
(Section~\ref{sec:exp-large-scale})
to examine performance at larger model scales
and across a broader range of mathematical reasoning benchmarks.

We use MaxRL~\citep{tajwar2026maxrl} as our primary
baseline to assess the performance gains of \method{}.
We match the total training rollout budget between
the two methods and use the same loss normalization
within each comparison.
On SmolLM2-360M-Instruct, we additionally compare against
GRPO~\citep{shao2024deepseekmath},
PKPO~\citep{walder2025pkpo}, and the rollout allocation
method DARS-HW~\citep{yang2025dars}.
We report Pass@1 and multi-sample Pass@$K$ to assess
single-sample accuracy and solution coverage, respectively.

\begin{figure}[t]
    \centering
    \includegraphics[width=\linewidth]
        {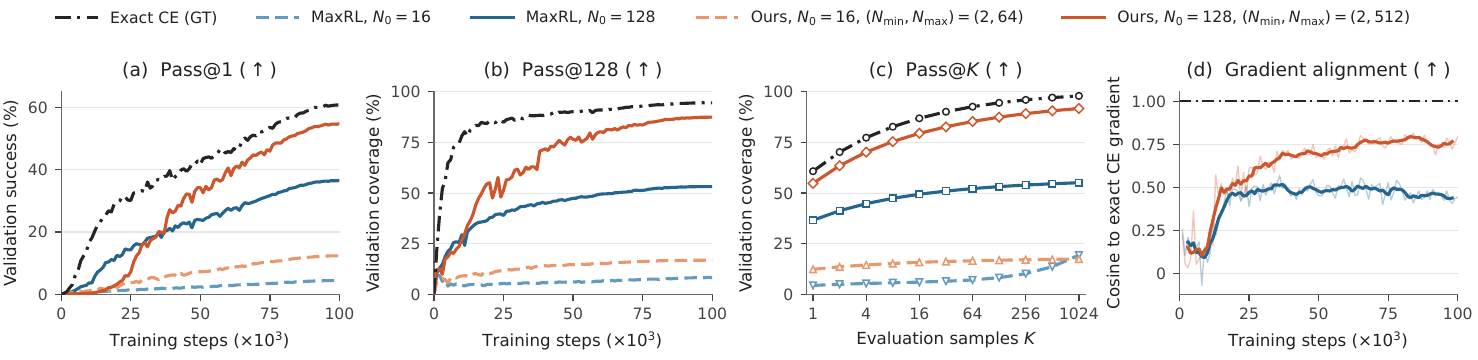}
    \caption{\textbf{Oracle rollout allocation on ImageNet.}
    MaxRL and \method{} (Ours) are compared under matched
    rollout budgets.
    (a,b) Validation Pass@1 and Pass@128 during training.
    (c) Pass@$K$ at step $100{,}000$.
    Black curves in (a--c) show the performance of
    a model trained directly with cross-entropy (CE).
    (d) Sampled-to-exact CE gradient cosine at $N_0=128$,
    measured along each method's own training trajectory.
    Faint and thick lines in (d) show raw measurements
    and centered five-point moving averages, respectively.}
    \label{fig:imagenet-oracle-allocation}
\end{figure}

\subsection{Exact-Likelihood Reference on ImageNet}
\label{sec:exp-imagenet}

We first evaluate how rollout allocation affects the fidelity of finite-rollout updates to the exact maximum-likelihood gradient.
Following MaxRL, we consider ImageNet classification
with a ResNet-50~\citep{he2016deepresidual} classifier.
Each rollout samples a class label and receives a
binary correctness reward.
In this setting, minimizing cross-entropy (CE) exactly
optimizes the log-success objective, providing a
reference without finite-rollout estimation error.
We compare uniform-rollout MaxRL and \method{} under
matched rollout budgets, alongside direct CE training.
Our allocation uses the exact success probabilities
available from the classifier, while policy updates
still rely on sampled labels and rewards.
This controlled setting isolates the effect of rollout allocation from success-probability estimation error.

Figures~\ref{fig:imagenet-oracle-allocation} (a,b)
show validation Pass@1 and Pass@128 during training.
With the larger rollout budget, \method{} progressively
separates from MaxRL and narrows the gap to direct
CE training in both metrics.
The advantage persists during later training,
demonstrating sustained improvements in both
single-sample performance and solution coverage.
With the smaller budget, \method{} also improves
Pass@1 and Pass@128. Figure~\ref{fig:imagenet-oracle-allocation} (c)
shows the final Pass@$K$ profiles.
At the larger training budget, \method{} outperforms
MaxRL across all displayed evaluation budgets,
extending the gains beyond Pass@1 and Pass@128.
At the smaller training budget, \method{} leads
over most of the displayed range.
The larger training budget thus produces the
broadest coverage gains.

Figure~\ref{fig:imagenet-oracle-allocation} (d)
compares the realized sampled training-loss gradient
with the exact CE gradient, computed at the same policy
parameters and from the same forward pass within each run.
The CE reference covers the full candidate batch.
At $N_0=128$, \method{} achieves a mean cosine similarity
of $0.6266$ between the sampled training-loss gradient and
the exact CE gradient, compared with $0.4317$ for MaxRL,
with the advantage persisting into later training.
These measurements provide direct gradient-level evidence
of stronger alignment with the likelihood gradient along
the observed training trajectories.
Together with the validation results, they show that
\method{} improves both sampled-gradient alignment and
held-out performance under matched rollout budgets. Further experimental details and additional results
are provided in Appendices~\ref{app:setup-imagenet}
and~\ref{app:results-imagenet}.

\begin{figure}[t]
    \centering
    \includegraphics[width=\linewidth]{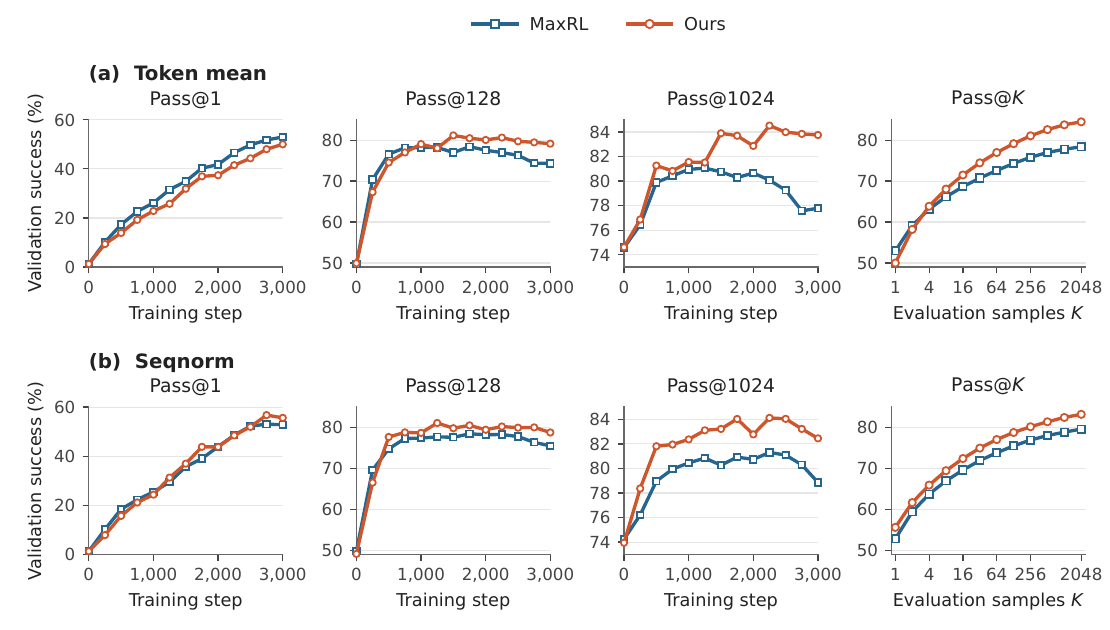}
    \caption{\textbf{Maze navigation.}
    MaxRL and \method{} (Ours) under (a) token-mean and
    (b) fixed-length sequence normalization (Seqnorm).
    Columns show Pass@1, Pass@128, and Pass@1024
    during training, followed by Pass@$K$ at step $3{,}000$.}
    \label{fig:exp-maze-dynamics}
\end{figure}

\subsection{Maze Navigation with a Lightweight Transformer}
\label{sec:exp-maze}

We next evaluate \method{} on maze navigation
with verifier-based rewards.
Following MaxRL, we use a
lightweight decoder-only Transformer based on the
Qwen2 architecture~\citep{yang2023qwen2}
to solve $17\times17$ mazes.
Given a maze, the model generates navigation actions
and receives a binary reward for reaching the goal
without violating the maze constraints.
We initialize RL training from the same pretrained
checkpoint used in MaxRL and train on a fixed
subset of mazes over multiple epochs,
evaluating on a separate held-out set.
We compare MaxRL and \method{} under
matched rollout budgets using both token-mean
normalization and fixed-length sequence normalization
(Seqnorm).
For rollout allocation, \method{} uses historical success-probability estimates from previous epochs, as described in Section~\ref{sec:practical-algorithm}.

Figure~\ref{fig:exp-maze-dynamics} shows Pass@1, Pass@128, and Pass@1024 during training,
alongside the final Pass@$K$ profiles.
Under both normalizations, solution coverage
initially improves for both methods.
During later training, MaxRL's high-budget coverage
declines, while \method{} mitigates this decline
and retains higher Pass@128 and Pass@1024. The final Pass@$K$ profiles provide a broader
assessment of maze-solving performance across
the evaluated range of sampling budgets.
Under token-mean normalization
(Figure~\ref{fig:exp-maze-dynamics} (a)),
\method{} achieves higher Pass@$K$ than MaxRL
at larger evaluation budgets, with the advantage
growing as $K$ increases.
Under Seqnorm
(Figure~\ref{fig:exp-maze-dynamics} (b)),
\method{} outperforms MaxRL across the entire
displayed Pass@$K$ profile. Together, these results demonstrate that our method
improves high-budget solution coverage under the
same rollout budget, extending the benefits
observed on ImageNet to sequence generation
with verifier-based rewards.
Further experimental details and additional results
are provided in Appendices~\ref{app:setup-maze} and~\ref{app:results-maze}.

\subsection{Mathematical Reasoning with SmolLM2}
\label{sec:exp-smollm}

We next evaluate \method{} on mathematical reasoning
with SmolLM2-360M-Instruct~\citep{allal2025smollm2}.
Following MaxRL, we train the model
on GSM8K~\citep{cobbe2021gsm8k} with binary rewards and evaluate on GSM8K-Platinum~\citep{vendrow2025reliability}.
We use a reference rollout count of $N_0=16$,
with \method{} allocating at most $N_{\max}=64$
rollouts per prompt.
We compare MaxRL and \method{} under matched
training rollout budgets using both token-mean
normalization and fixed-length sequence normalization
(Seqnorm).

Table~\ref{tab:exp-smollm-results} reports
Pass@1, Pass@32, Pass@128, and Pass@256
at training step $2{,}000$, with the initial model included as a pre-RL reference.
Under both normalizations, \method{} improves
Pass@32, Pass@128, and Pass@256 over MaxRL,
with little change in Pass@1.
The consistent gains across these evaluation budgets
demonstrate improved solution coverage from
redistributing the same number of training responses. We further compare \method{} with additional
RL and rollout allocation baselines.
\method{} outperforms GRPO~\citep{shao2024deepseekmath}
at every reported $K$.
PKPO~\citep{walder2025pkpo} achieves higher
Pass@$K$ at large $K$, but substantially
lower Pass@1 than \method{}.
We also include DARS-HW~\citep{yang2025dars}
as a closely related allocation baseline.
Our allocation is derived specifically from
MaxRL's finite-sample gradient scaling and
enforces a fixed total rollout budget,
whereas DARS-HW imposes per-prompt limits
without fixing the batch-level total.
In this comparison, \method{} achieves higher
Pass@32, Pass@128, and Pass@256 than DARS-HW
using fewer training responses.
DARS-HW attains higher Pass@1 and consumes
approximately $1.49\times$ as many training
rollouts as other methods over the same number
of training steps. These results demonstrate the effectiveness of
\method{} in improving mathematical reasoning
coverage under a fixed training rollout budget.
Further experimental details and additional results
are provided in Appendices~\ref{app:setup-smollm} and~\ref{app:results-smollm}.

\begin{table}[t]
    \centering
    \caption{\textbf{SmolLM2-360M-Instruct on GSM8K-Platinum.}
    All RL methods are evaluated at step $2{,}000$,
    with the base model as the pre-RL reference.
    Values are percentages; $\Delta$ reports percentage-point
    changes relative to MaxRL under the same normalization.
    $\dagger$: DARS-HW uses $1.49\times$ as many training
    rollouts as the other methods.}
    \label{tab:exp-smollm-results}
    \begin{tabular}{lrrrr}
        \toprule
        Method & Pass@1 & Pass@32 & Pass@128 & Pass@256 \\
        \midrule
        \textcolor{black!60}{Base model}
            & \textcolor{black!60}{6.80}
            & \textcolor{black!60}{53.85}
            & \textcolor{black!60}{76.32}
            & \textcolor{black!60}{84.37} \\
        \midrule
        GRPO~\citep{shao2024deepseekmath}
            & 30.46 & 45.80 & 49.28 & 50.62 \\
        PKPO (T=16)~\citep{walder2025pkpo}
            & 19.93 & 70.32 & 84.92 & 90.16 \\
        MaxRL (token-mean)~\citep{tajwar2026maxrl}
            & 32.60 & 59.41 & 68.44 & 72.21 \\
        MaxRL (Seqnorm)~\citep{tajwar2026maxrl}
            & 32.08 & 56.66 & 64.58 & 68.24 \\
        \midrule
        DARS-HW$^\dagger$~\citep{yang2025dars}
            & 34.25 & 52.38 & 57.31 & 59.06 \\
        \method{} (token-mean)
            & 33.01 & 63.38 & 73.17 & 77.34 \\
        \quad $\Delta$ vs.\ MaxRL (token-mean)
            & \textcolor{green!50!black!60}{$+0.41$}
            & \textcolor{green!50!black!60}{$+3.97$}
            & \textcolor{green!50!black!60}{$+4.73$}
            & \textcolor{green!50!black!60}{$+5.13$} \\
        \addlinespace[2pt]
        \method{} (Seqnorm)
            & 32.00 & 61.35 & 70.12 & 73.37 \\
        \quad $\Delta$ vs.\ MaxRL (Seqnorm)
            & \textcolor{red!70!black!60}{$-0.08$}
            & \textcolor{green!50!black!60}{$+4.69$}
            & \textcolor{green!50!black!60}{$+5.54$}
            & \textcolor{green!50!black!60}{$+5.13$} \\
        \bottomrule
    \end{tabular}
\end{table}

\subsection{Mathematical Reasoning with Qwen}
\label{sec:exp-large-scale}

We further evaluate \method{} on mathematical reasoning
with Qwen2.5-Math-1.5B~\citep{yang2024qwen25math}
and Qwen3-4B-Base~\citep{yang2025qwen3}.
For Qwen2.5-Math-1.5B, we train on MATH level 3--5
problems~\citep{hendrycks2021math}. Evaluation covers BeyondAIME~\citep{bytedanceseed2025beyondaime},
AIME 2025~\citep{zhang2025aime25},
MATH-500~\citep{lightman2024verify},
and OlympiadBench~\citep{he2024olympiadbench}.
For Qwen3-4B-Base, we train on
POLARIS-53K~\citep{an2025polaris}, following the MaxRL training setup. We evaluate our checkpoint at step $1{,}000$ against
the publicly released GRPO and MaxRL checkpoints
at the same training step~\citep{tajwar2026maxrl}.
All three checkpoints are evaluated using the same
sampling and scoring protocol on BeyondAIME,
AIME 2025, MATH-500, and Minerva Math~\citep{lewkowycz2022minerva}.

\begin{figure}[t]
    \centering
    \includegraphics[width=\linewidth]{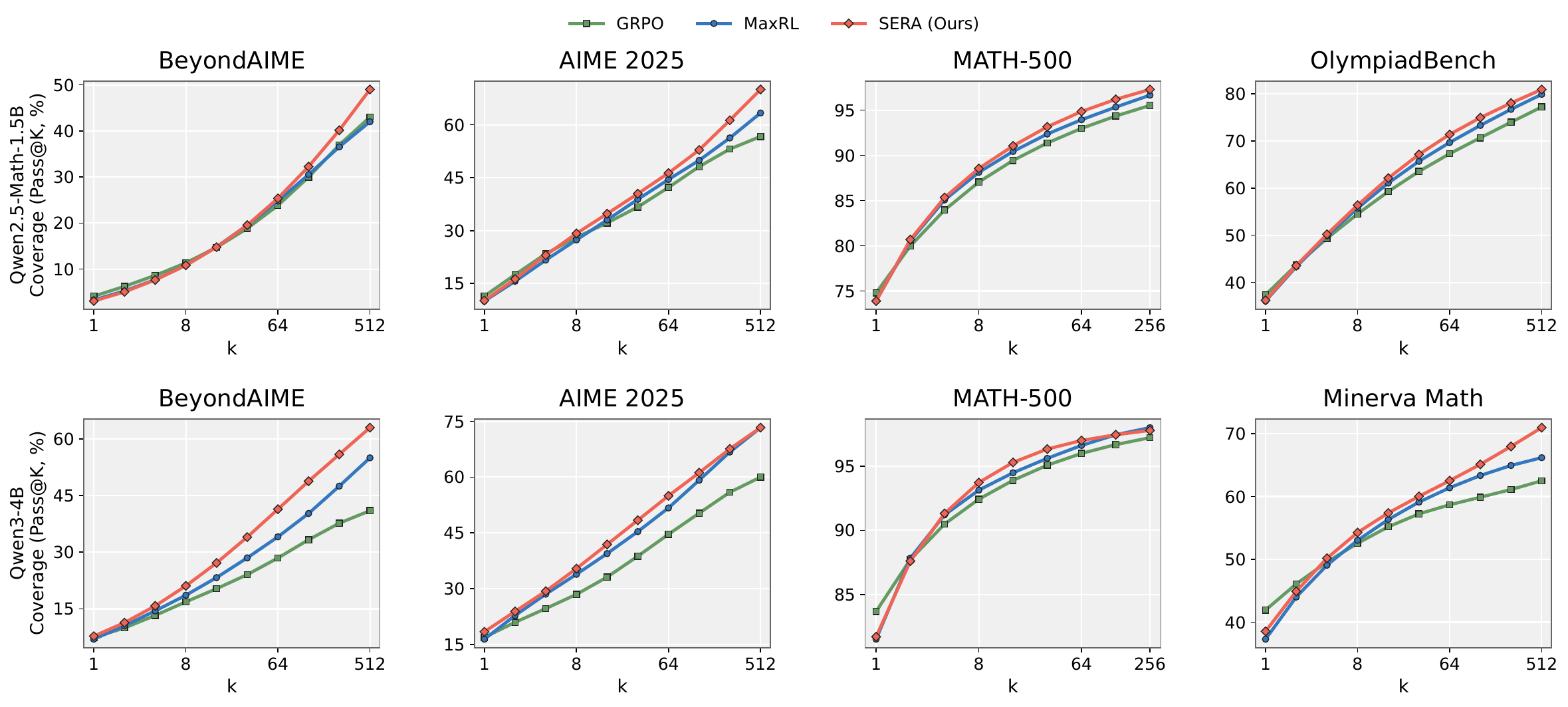}
    \caption{\textbf{Pass@$K$ profiles for Qwen models.}
    Top: Qwen2.5-Math-1.5B at step $600$.
    Bottom: Qwen3-4B-Base at step $1{,}000$.
    Both rows compare GRPO, MaxRL, and \method{}.}
    \label{fig:exp-qwen}
\end{figure}

Figure~\ref{fig:exp-qwen} compares the Pass@$K$ profiles
of Qwen2.5-Math-1.5B and Qwen3-4B.
Across both model scales, \method{} achieves higher Pass@$K$
than GRPO at the largest displayed $K$ on all eight
model-benchmark pairs.
Relative to MaxRL, \method{} consistently improves the
high-budget coverage of Qwen2.5-Math-1.5B, attaining higher
Pass@$K$ at every reported $K\ge32$ on all four benchmarks
while keeping Pass@1 within $0.22$ percentage points.
The gains increase with the evaluation budget on BeyondAIME
and AIME 2025, reaching $7.00$ and $6.67$ percentage points
at $K=512$, respectively.
On Qwen3-4B, \method{} improves Pass@1 on all four benchmarks
and delivers substantial Pass@512 gains of $8.00$ points on
BeyondAIME and $4.78$ points on Minerva Math.
It matches MaxRL at $K=512$ on AIME 2025.
On MATH-500, MaxRL and \method{} already achieve approximately
$97$--$98\%$ coverage at $K=256$ across both model scales,
leaving limited headroom for absolute improvements. Overall, these results show that \method{} transfers across
model scales and training datasets, improving the
single-sample to multi-sample performance profile under
matched training rollout budgets, with the largest gains
on benchmarks that retain substantial high-budget headroom.
Further experimental details and additional results
are provided in Appendices~\ref{app:setup-qwen}
and~\ref{app:results-qwen}.

\section{Related Work}
\label{sec:related-work}

\paragraph{RL objectives and reasoning coverage.}
GRPO~\citep{shao2024deepseekmath} and
DeepSeek-R1~\citep{guo2025deepseekr1} established group-based
reinforcement learning with verifiable rewards as an effective
paradigm for reasoning models. Subsequent work studies how the
objective and its estimator shape the update, including normalization
effects in Dr.~GRPO~\citep{liu2025understanding}, sequence-level
ratios in GSPO~\citep{zheng2025gspo}, and transformed
success-probability objectives~\citep{davis2025objective}.
These choices matter for repeated sampling: Pass@$K$ measures
multi-sample solution coverage, and improvements in Pass@1 need not
improve coverage at larger $K$~\citep{chen2021evaluating,
yue2025reasoningcapacity,barakat2026promptinterference}.
Inference-aware objectives such as PKPO~\citep{walder2025pkpo}
and diversity-oriented objectives such as SetPO~\citep{li2026setpo}
and UCPO~\citep{lochab2026ucpo} explicitly target multi-sample or diversity-sensitive behavior. MaxRL~\citep{tajwar2026maxrl}, in contrast, targets prompt-wise log-success using a centered finite-rollout estimator. \method{} keeps the same log-success target and instead reallocates a fixed rollout budget to control the prompt-dependent finite-rollout scaling of the centered estimator.

\paragraph{Adaptive rollout allocation and selection.}
Recent methods allocate or filter rollouts using different notions of
value: Knapsack RL~\citep{li2025knapsack} uses estimated learning
gain, VIP~\citep{nguyen2026vip} and DynaMO~\citep{fang2026dynamo}
use variance-related signals, and HORA~\citep{wang2026hora}
optimizes the probability of finding successful responses. DARS-HW
\citep{yang2025dars} gives a likelihood-oriented inverse-success
schedule, while Reinforce-Ada~\citep{xiong2025reinforceada}
derives adaptive sampling and weighting from nonlinear objectives.
Other work makes decisions through selection, replay, or sequential
continuation, including DAPO~\citep{yu2025dapo},
Pilot-Commit~\citep{kim2026pilotcommit}, and SARA~\citep{nomand2026sara}.
\method{} differs in the quantity it controls: for a fixed retained
set, it solves a bounded max--min problem over the finite-rollout
scaling coefficients of the centered MaxRL estimator under a fixed
total rollout budget. Historical estimates determine the inputs to
this allocation, and multiplicity correction removes the additional
response-count weighting induced by unequal group sizes. Appendix~\ref{app:extended-related-work}
gives the detailed comparison.

\section{Conclusion}

We considered the problem of realizing MaxRL's likelihood objective under a fixed rollout budget.
Our theoretical analysis shows that uniform rollout counts induce unequal
attenuation of prompt-wise likelihood gradients, more strongly attenuating gradients from low-success prompts and potentially
shifting the expected update direction.
Building on this analysis, we proposed \method{}, which
redistributes the available rollouts to approximately equalize
finite-rollout gradient scaling across retained prompts while correcting the additional prompt weighting introduced
by unequal rollout counts.
Experiments show stronger alignment with exact likelihood
gradients on ImageNet and improved multi-sample solution
coverage over MaxRL on maze navigation and mathematical
reasoning under matched training rollout budgets. 
On the mathematical reasoning benchmarks, Pass@1 remains comparable or improves while multi-sample coverage increases.
Overall, \method{} provides a principled way to redistribute a fixed rollout budget so as to better preserve MaxRL's intended gradient structure while improving downstream multi-sample coverage.

\subsection*{AI use statement}

Generative AI tools were used to assist with language editing and to
improve the clarity and readability of the paper. All AI-assisted edits were reviewed and verified
by the authors. The authors take full responsibility for the final content
of this paper.

\subsection*{Reproducibility statement}

We provide the information required to reproduce the theoretical and
empirical results of this work. The complete derivations and proofs of the
theoretical results are provided in the appendix. Detailed experimental
settings, including datasets, model architectures, training configurations,
optimization hyperparameters, rollout budgets, evaluation protocols, and
computational resources, are reported in the Experimental Details appendix.
Additional results and training diagnostics are also provided in the
appendix. Our implementation follows the algorithm described in the main
paper, and the code necessary to reproduce the experiments is provided
as supplementary material.

\bibliography{iclr2027_conference}
\bibliographystyle{iclr2027_conference}

\clearpage

\appendix

\section*{Contents of the Appendix}

\makeatletter
\begingroup
\c@tocdepth=3\relax

\newcommand{\appTOCentry}[3]{%
  \@dottedtocline{1}{#1}{3em}{%
    \hyperref[#2]{\numberline{\ref*{#2}}#3}%
  }{\pageref{#2}}%
}

\appTOCentry{0em}{app:extended-related-work}
  {\textbf{Extended Related Work}}

\medskip

\appTOCentry{0em}{app:additional-theory}
  {\textbf{Additional Theory}}

\appTOCentry{1.5em}{app:prelim-details}
  {Finite-Rollout Fidelity of the Centered Estimator}

\appTOCentry{1.5em}{app:heterogeneous-objective}
  {Conditional Objectives and Batch-Gradient Direction}

\appTOCentry{1.5em}{app:waterline}
  {Derivation of the Fixed-Budget Allocation}

\appTOCentry{1.5em}{app:long-tail-observability}
  {Mixed-Group Observability and Budget Concentration}

\appTOCentry{1.5em}{app:inverse-fidelity}
  {Comparison with Inverse-Fidelity Reweighting}

\medskip

\appTOCentry{0em}{app:implementation-details}
  {\textbf{Implementation Details}}

\appTOCentry{1.5em}{app:p-estimation}
  {Historical Success-Probability Estimates}

\appTOCentry{1.5em}{app:integerization}
  {Bounded Allocation and Exact-Budget Integerization}

\appTOCentry{1.5em}{app:active-set-details}
  {Active-Prefix Search and Capacity Fallback}

\appTOCentry{1.5em}{app:loss-normalization}
  {Loss Aggregation and Multiplicity Correction}

\medskip

\appTOCentry{0em}{app:experimental-details}
  {\textbf{Extended Experimental Details}}

\appTOCentry{1.5em}{app:experimental-setup}
  {Experimental Setup}

\appTOCentry{3em}{app:setup-imagenet}
  {ImageNet}

\appTOCentry{3em}{app:setup-maze}
  {Maze}

\appTOCentry{3em}{app:setup-smollm}
  {SmolLM2 on GSM8K}

\appTOCentry{3em}{app:setup-qwen}
  {Qwen}

\appTOCentry{1.5em}{app:more-results}
  {More Experimental Results}

\appTOCentry{3em}{app:results-imagenet}
  {ImageNet}

\appTOCentry{3em}{app:results-maze}
  {Maze}

\appTOCentry{3em}{app:results-smollm}
  {SmolLM2 on GSM8K}

\appTOCentry{3em}{app:results-qwen}
  {Mathematical reasoning with Qwen}

\appTOCentry{3em}{app:training-dynamics}
  {Training Dynamics and Allocation Behavior}

\endgroup
\makeatother

\section{Extended Related Work}
\label{app:extended-related-work}

\paragraph{Reinforcement learning for reasoning.}
Language-model post-training includes preference-based optimization with
PPO~\citep{schulman2017ppo,ouyang2022training} and REINFORCE-style
methods without a learned value function~\citep{ahmadian2024rloo}.
For reasoning with verifiable rewards, GRPO~\citep{shao2024deepseekmath}
constructs advantages from sampled response groups, and
DeepSeek-R1~\citep{guo2025deepseekr1} demonstrates the effectiveness
of this paradigm at scale.
The update depends on both the objective and its gradient estimator:
Dr.~GRPO~\citep{liu2025understanding} analyzes biases introduced by
loss and advantage normalization, while GSPO~\citep{zheng2025gspo}
uses sequence-level importance ratios and clipping.
\citet{davis2025objective} further interprets binary-reward updates
through transformed success-probability objectives.

Multi-sample evaluation introduces an additional distinction between
single-sample accuracy and solution coverage.
Pass@$K$ estimates the probability that at least one of $K$ sampled
responses is correct~\citep{chen2021evaluating}.
RL can improve Pass@1 while reducing coverage at larger sampling
budgets~\citep{yue2025reasoningcapacity}; analyses of prompt
interference and training objectives help explain why gains across
evaluation budgets need not coincide~\citep{barakat2026promptinterference,yu2025passkdiagnostic}.
\citet{tang2025inference} formulate objectives for inference-time
aggregation, including Pass@$K$ and majority voting.
PKPO~\citep{walder2025pkpo} and Pass@$K$
training~\citep{chen2025passktraining} target multi-sample success,
while max@$K$ optimization~\citep{bagirov2025maxk} targets expected
maximum reward.

Another line of work explicitly promotes diversity among solutions.
SetPO~\citep{li2026setpo} incorporates each trajectory's contribution
to set-level diversity, UCPO~\citep{lochab2026ucpo} regularizes the
conditional distribution over correct responses, and
VPO~\citep{bahlousboldi2026vpo} encourages specialization across
reward scalarizations.
These training interventions are distinct from inference-time answer
selection through self-consistency or majority
voting~\citep{wang2023selfconsistency,lewkowycz2022minerva}, or
through learned outcome and process
verifiers~\citep{cobbe2021gsm8k,lightman2024verify}.

MaxRL~\citep{tajwar2026maxrl} develops compute-indexed approximations
to prompt-wise log-success maximization and characterizes the
finite-rollout behavior of its practical centered estimator.
\method{} adopts this likelihood target and studies how rollout
allocation controls the relative scaling of prompt gradients under a
fixed rollout budget.
Pass@$K$ profiles assess the resulting solution coverage across
inference budgets, while the allocation objective concerns the
finite-rollout approximation to the log-success gradient.

\paragraph{Adaptive rollout allocation.}
Adaptive allocation methods differ in how they value additional
responses, which prompts and responses they retain, and when they
make sampling decisions.
\citet{liao2025efficiency} combine difficulty-based allocation with
temperature scheduling.
Knapsack RL~\citep{li2025knapsack} combines mixed-reward probability
with estimated learning gain.
GVM-RAFT~\citep{yao2025gvm} studies variance-based allocation for
rejection sampling and extends its framework to GRPO, while
VIP~\citep{nguyen2026vip} predicts success probabilities to minimize
expected gradient variance.
DynaMO~\citep{fang2026dynamo} uses historical Bernoulli reward
variance as an allocation proxy alongside token-level advantage
modulation.
HORA~\citep{wang2026hora} maximizes the sum of posterior hit
probabilities for additional rollouts, and
CERO~\citep{zong2026cero} uses concave utilities of cumulative
budgets across epochs.
These criteria emphasize learning gain, variance reduction,
successful-response discovery, or utility over a training horizon.

Selection methods act at different stages of generation and optimization.
DAPO~\citep{yu2025dapo} filters identical-reward groups and continues
sampling to collect informative prompts.
GRESO~\citep{zheng2025greso} uses historical outcomes to skip likely
uninformative prompts before generation, while
DOTS~\citep{sun2025dots} combines difficulty-targeted selection with
rollout replay.
POLARIS~\citep{an2025polaris} reuses successful responses to rescue
all-failure groups.
PODS~\citep{xu2025pods} instead selects a subset of already generated
responses to reduce policy-update cost.
These approaches distinguish avoiding generation, reusing responses,
and reducing the update batch.

Success-probability estimation for rollout scheduling commonly uses
pilot samples, historical statistics, or predictive models.
Pilot-based methods estimate prompt difficulty from a small set of
fresh rollouts before allocating additional rollouts, as in
DARS~\citep{yang2025dars}.
History-based methods reuse outcomes from earlier visits or epochs:
Knapsack RL~\citep{li2025knapsack} uses success rates observed in
the previous epoch, while the Ada-EMA variant of
Reinforce-Ada~\citep{xiong2025reinforceada} maintains discounted
success and trial counts with Bayesian smoothing.
MoPPS~\citep{qu2025mopps} also uses streaming Bayesian estimates
to track prompt success rates.
Predictive models additionally exploit relationships among prompts:
VIP~\citep{nguyen2026vip} uses a Gaussian process over prompt
embeddings with recent rollout feedback, while graph-structured
estimation~\citep{liu2026graphscheduling} shares observations across
related prompts.
These approaches supply the estimates $\widehat p_q$;
the allocation criterion determines how those estimates translate
into rollout counts.

Sequential allocation additionally uses outcomes observed during
rollout collection.
Pilot-Commit~\citep{kim2026pilotcommit} screens prompts with pilot
rollouts before committing further budget.
VIGOR~\citep{jiang2026vigor} progressively expands sampling for
prompts with high observed reward variance, while
SARA~\citep{nomand2026sara} uses sequential evidence to commit or
abandon groups.
DATPO~\citep{yu2026datpo} adapts generation through
difficulty-dependent tree budgets and entropy-based forking.
Adaptive continuation also changes which gradient terms become
observable.
PAIR~\citep{nomand2026pair} uses inverse joint inclusion
probabilities to recover the complete-candidate pair gradient in
expectation for an unclipped, unstandardized leave-one-out estimator
under its stated sampling assumptions.

The DARS-HW variant of DARS~\citep{yang2025dars} and
Reinforce-Ada~\citep{xiong2025reinforceada} are particularly close
to \method{} through their likelihood-oriented formulations.
DARS uses preliminary outcomes to allocate further rollouts to
difficult prompts; its HW schedule is the variant whose idealized
prompt-level target is maximum likelihood.
Under the population-baseline approximation used in its analysis,
the idealized schedule $N_q\propto 1/p_q$ makes the summed
reward-gradient contribution $N_q\nabla_\theta p_q$ proportional
to $\nabla_\theta\log p_q$.
Reinforce-Ada derives the weight $f'(p_q)$ from a nonlinear objective
$f(p_q)$ and realizes it through sampling, explicit gradient
weighting, or a combination of the two, including $f(p)=\log p$.
These likelihood-oriented formulations share MaxRL's log-success
target.
The comparison with \method{} therefore concerns the estimator
being controlled and the role assigned to rollout allocation.

\method{} starts from the practical centered MaxRL estimator in
Equation~\ref{eq:practical-maxrl}.
Let $g_q=\nabla_\theta\log p_q$ and
$\kappa_q=1-(1-p_q)^{N_q-1}$, so that
$\mathbb E[\widehat g_q^{(N_q)}]=\kappa_q g_q$ for a fixed count.
Consider a candidate batch of size $B$, a retained set $\mathcal A$,
and a total rollout budget $M=BN_0$.
Let $\mathcal F$ denote the current policy, candidate batch,
historical information, and allocation decisions fixed before fresh
on-policy sampling.
Under response-mean sequence-score aggregation, the conditional
means without and with the per-response weight $N_0/N_q$ are
\begin{equation}
\mathbb E[\widehat G_{\mathrm{plain}}\mid\mathcal F]
=\frac{1}{B}\sum_{q\in\mathcal A}
  \frac{N_q}{N_0}\kappa_q g_q,
\qquad
\mathbb E[\widehat G_{\mathrm{corr}}\mid\mathcal F]
=\frac{1}{B}\sum_{q\in\mathcal A}\kappa_q g_q.
\label{eq:rw-multiplicity}
\end{equation}
The correction removes the additional response-count weighting,
while allocation controls the remaining finite-rollout coefficients
(Appendix~\ref{app:loss-normalization}).
For a fixed retained set and frozen success estimates, the bounded
continuous allocator maximizes
$\min_{q\in\mathcal A}\kappa(\widehat p_q,x_q)$ subject to
$\sum_{q\in\mathcal A}x_q=M$ and
$N_{\min}\leq x_q\leq N_{\max}$.
Its interior waterline rule satisfies
$x_q-1= c_{\mathcal B}/[-\log(1-\widehat p_q)]$
(Appendix~\ref{app:waterline}), with inverse-success behavior
arising as a small-$\widehat p_q$ approximation.
The practical procedure converts these counts to integers while
preserving the budget and bounds.
Thus, \method{} allocates samples to control the finite-rollout
scaling of an already likelihood-oriented estimator and explicitly
removes the additional weighting caused by unequal group sizes.
Together, these components target comparable scaling of prompts' log-success gradients under a fixed rollout budget
(Appendix~\ref{app:batch-fidelity}).

\section{Additional Theory}
\label{app:additional-theory}

\subsection{Finite-Rollout Fidelity of the Centered Estimator}
\label{app:prelim-details}
\label{app:n-minus-one}

Following the finite-rollout analysis of
MaxRL~\citep{tajwar2026maxrl}, we derive the expectation of
the centered sequence-score estimator used in our analysis.
Fix a prompt $q$ and draw $N\ge1$ i.i.d.\ responses from
$\pi_\theta(\cdot\mid q)$, with $N$ chosen before sampling.
The verifier is binary and does not explicitly depend on
$\theta$. Suppressing the prompt subscript, write
$p=\mathbb E[R]\in(0,1)$,
$s=\nabla_\theta\log\pi_\theta(y\mid q)$, and
$g=\nabla_\theta\log p$.
Under the usual score-function regularity conditions,
\begin{equation}
    \mathbb E[s]=0,
    \qquad
    \mathbb E[Rs]=\nabla_\theta p=pg.
    \label{eq:app-score-identities}
\end{equation}
It follows that
\begin{equation}
    \mu_1:=\mathbb E[s\mid R=1]=g,
    \qquad
    \mu_0:=\mathbb E[s\mid R=0]
    =-\frac{p}{1-p}g.
    \label{eq:conditional-scores}
\end{equation}

\paragraph{Finite-rollout expectation.}
The centered estimator in
Equation~\ref{eq:practical-maxrl} satisfies
\begin{equation}
    \mathbb E[\widehat g^{(N)}]
    =\kappa(p,N)g,
    \qquad
    \kappa(p,N):=1-(1-p)^{N-1}.
    \label{eq:appendix-fidelity-identity}
\end{equation}

\paragraph{Proof.}
Let $C:=\sum_{i=1}^{N}R_i
\sim\operatorname{Binomial}(N,p)$.
For $C>0$, the estimator can be written as
\begin{equation}
    \widehat g^{(N)}
    =\frac1C\sum_{i=1}^{N}R_i s_i
     -\frac1N\sum_{i=1}^{N}s_i.
    \label{eq:practical-maxrl-decomposition}
\end{equation}
Conditional on $C=k>0$, the successful and unsuccessful
responses have expected scores $\mu_1$ and $\mu_0$,
respectively. Hence,
\begin{align}
    \mathbb E[\widehat g^{(N)}\mid C=k]
    &=\mu_1-\frac{k\mu_1+(N-k)\mu_0}{N}
      \nonumber\\
    &=\frac{N-k}{N(1-p)}g.
    \label{eq:app-centered-given-count}
\end{align}
Since the estimator is zero when $C=0$,
\begin{align}
    \mathbb E[\widehat g^{(N)}]
    &=\frac{
        \mathbb E[(1-C/N)\mathbf1_{\{C>0\}}]
      }{1-p}g
      \nonumber\\
    &=\frac{1-p-(1-p)^N}{1-p}g
      \nonumber\\
    &=[1-(1-p)^{N-1}]g.
\end{align}

\subsection{Conditional Objectives and Batch-Gradient Direction}
\label{app:heterogeneous-objective}
\label{app:batch-fidelity}

Let $\mathcal B$ be a candidate batch with
$B:=|\mathcal B|$, and let
$\mathcal A\subseteq\mathcal B$ be a nonempty retained set.
We condition on the history, current policy, candidate batch,
and the set and integer counts chosen before the current
rollouts; denote this information by $\mathcal F$.
The set $\mathcal A$ and counts $N_q$ are held fixed
when differentiating with respect to $\theta$.
Write $g_q:=\nabla_\theta\log p_q$ and
$\kappa_q:=\kappa(p_q,N_q)$.

\paragraph{A conditional objective for heterogeneous counts.}
For an integer $m\ge0$, define
\begin{equation}
    \Phi_m(p):=-\sum_{j=1}^{m}\frac{(1-p)^j}{j},
    \qquad
    \Phi_0(p):=0.
    \label{eq:app-truncated-log-objective}
\end{equation}
Differentiation and the log-series identity give
\begin{equation}
    \Phi_m'(p)=\frac{1-(1-p)^m}{p},
    \qquad
    \lim_{m\to\infty}\Phi_m(p)=\log p.
\end{equation}
For the current retained set and counts, define
\begin{equation}
    J_{\mathcal A,\mathbf N}(\theta)
    :=\frac1B\sum_{q\in\mathcal A}
      \Phi_{N_q-1}(p_q(\theta)).
    \label{eq:heterogeneous-surrogate-objective}
\end{equation}
The finite-rollout identity then implies
\begin{align}
    \nabla_\theta J_{\mathcal A,\mathbf N}(\theta)
    &=\frac1B\sum_{q\in\mathcal A}\kappa_q g_q
      \nonumber\\
    &=\mathbb E[\widehat G\mid\mathcal F],
    \qquad
    \widehat G:=
      \frac1B\sum_{q\in\mathcal A}\widehat g_q^{(N_q)}.
    \label{eq:app-conditional-batch-objective}
\end{align}
Heterogeneous counts therefore assign different truncation
orders of the same log-success expansion to different prompts.
This characterizes the expected on-policy estimator at the
current allocation.

\paragraph{Relative scaling and direction.}
Using the original batch normalizer, define the retained-set
maximum-likelihood reference and the conditional mean estimator:
\begin{equation}
    G_{\mathcal A}^{\log}
    :=\frac1B\sum_{q\in\mathcal A}g_q,
    \qquad
    \overline G:=\mathbb E[\widehat G\mid\mathcal F].
\end{equation}
For any $\gamma>0$,
\begin{equation}
    \overline G
    =\gamma G_{\mathcal A}^{\log}
     +\frac1B\sum_{q\in\mathcal A}
       (\kappa_q-\gamma)g_q.
    \label{eq:batch-direction-decomposition}
\end{equation}
If every retained prompt has $\kappa_q=\gamma$, then
$\overline G=\gamma G_{\mathcal A}^{\log}$.
Whenever $G_{\mathcal A}^{\log}\ne0$, the two vectors have
the same direction.
Unequal coefficients can alter the aggregate direction:
for two linearly independent prompt gradients, different
coefficients change their relative contributions to the sum.

\paragraph{Approximate equalization.}
Let
$\delta:=\max_{q\in\mathcal A}|\kappa_q-\gamma|$.
Applying the triangle inequality to
Equation~\ref{eq:batch-direction-decomposition} gives
\begin{equation}
    \left\|
        \overline G-\gamma G_{\mathcal A}^{\log}
    \right\|
    \le
    \frac{\delta}{B}
    \sum_{q\in\mathcal A}\|g_q\|.
    \label{eq:app-retained-equalization-bound}
\end{equation}
For fixed prompt gradients and reference scale, reducing
the maximum coefficient deviation tightens this bound on
the conditional mean-gradient discrepancy.
Exact equalization makes the discrepancy zero.
These results motivate controlling the relative
finite-rollout coefficients across retained prompts.
SERA uses estimated success probabilities to target
this equalization.

\subsection{Derivation of the Fixed-Budget Allocation}
\label{app:waterline}

We derive the allocation in
Equations~\ref{eq:closed-form-allocation}
and~\ref{eq:shared-allocation-constant}
from the fixed-budget objective in
Equation~\ref{eq:waterline-objective}.
Consider a batch $\mathcal B$ of $B$ prompts with total
rollout budget $M=BN_0\ge B$.
We first treat the counts as continuous variables
$N_q\ge1$, before imposing practical count bounds
and integer constraints.
For $p_q\in(0,1)$, define
$h_q:=-\log(1-p_q)>0$, so that
\begin{equation}
    \kappa(p_q,N_q)
    =1-\exp[-h_q(N_q-1)].
    \label{eq:app-fidelity-exposure}
\end{equation}

\paragraph{Rollout counts required by a common fidelity target.}
Suppose every prompt must attain a fidelity of at least
$\gamma\in[0,1)$.
Writing $c:=-\log(1-\gamma)\ge0$, this requirement is
equivalent to
\begin{equation}
    \kappa(p_q,N_q)\ge\gamma
    \quad\Longleftrightarrow\quad
    h_q(N_q-1)\ge c
    \quad\Longleftrightarrow\quad
    N_q\ge1+\frac{c}{h_q}.
    \label{eq:app-target-count-requirement}
\end{equation}
Thus, a common fidelity target determines a minimum
rollout count for each prompt.

\paragraph{Determining the target from the budget.}
Summing the required counts over the batch gives
\begin{equation}
    M=\sum_{q\in\mathcal B}N_q
    \ge
    B+c\sum_{q\in\mathcal B}h_q^{-1}.
    \label{eq:app-target-budget-requirement}
\end{equation}
Consequently, every feasible target must satisfy
\begin{equation}
    c\le
    c_{\mathcal B}
    :=
    \frac{M-B}
         {\displaystyle\sum_{q\in\mathcal B}h_q^{-1}}.
    \label{eq:app-budget-determined-waterlevel}
\end{equation}
This upper bound is attained by setting
$N_q=1+c_{\mathcal B}/h_q$ for every prompt.
These counts sum to $M$ and give every prompt the same
fidelity $1-\exp(-c_{\mathcal B})$.
Any larger common fidelity would require more than $M$
rollouts, so this allocation maximizes the minimum
fidelity under the continuous budget constraint.

Substituting the budget-determined value of
$c_{\mathcal B}$ yields the explicit allocation
\begin{equation}
    N_q^\star
    =
    1+\frac{c_{\mathcal B}}{h_q}
    =
    1+(M-B)
      \frac{[-\log(1-p_q)]^{-1}}
           {\displaystyle
            \sum_{r\in\mathcal B}[-\log(1-p_r)]^{-1}}.
    \label{eq:app-explicit-budget-allocation}
\end{equation}
The factor $M-B$ follows from
$\sum_{q\in\mathcal B}(N_q-1)=M-B$:
the centered estimator's fidelity depends on $N_q-1$.
Lower success probabilities produce larger inverse
exposures $h_q^{-1}$ and therefore receive more rollouts.

\paragraph{Allocation on a retained set.}
The same derivation applies to a fixed nonempty retained set
$\mathcal A\subseteq\mathcal B$ of size $L=|\mathcal A|$.
Keeping the original total budget $M=BN_0$, we obtain
\begin{equation}
    N_q^\star
    =
    1+(M-L)
      \frac{[-\log(1-p_q)]^{-1}}
           {\displaystyle
            \sum_{r\in\mathcal A}[-\log(1-p_r)]^{-1}},
    \qquad q\in\mathcal A.
    \label{eq:unbounded-centered-waterline}
\end{equation}
Here, $\sum_{q\in\mathcal A}(N_q^\star-1)=M-L$.
The practical procedure uses frozen success estimates
$\widehat p_q$, incorporates count bounds, and converts
the continuous counts to integers while preserving the
total budget, as detailed in
Appendix~\ref{app:integerization}.

\paragraph{Optimality with count bounds.}
Assume $LN_{\min}\le M\le LN_{\max}$.
A common target $\kappa(p_q,N_q)\ge1-e^{-t}$,
with $t\ge0$, is feasible if and only if
\[
    t\le\min_{q\in\mathcal A}h_q(N_{\max}-1),
    \qquad
    \sum_{q\in\mathcal A}
    \max\!\left\{N_{\min},1+\frac{t}{h_q}\right\}\le M.
\]
The clipped waterline attains the largest feasible target
and therefore solves the bounded continuous max--min problem.

\subsection{Mixed-Group Observability and Budget Concentration}
\label{app:long-tail-observability}

\paragraph{Observability as a safeguard.}
The monotonicity of $U$ implies that
$U(p,N_{\max})<u_0$ rules out meeting the observability
threshold with any permitted rollout count.
Conversely, passing this maximum-count screen does not
ensure that a smaller allocated count meets the threshold.
SERA therefore evaluates observability again at the final
integer counts, using the frozen success estimates
$\widehat p_q$.

The quantities $\kappa$ and $U$ describe different aspects
of the estimator. At fixed $N\ge2$, $\kappa(p,N)$ increases
with $p$, whereas $U(p,N)$ is symmetric around $p=1/2$.
In particular, as $p\to1$, $\kappa(p,N)\to1$ while
$U(p,N)\to0$.
The former describes the multiplicative coefficient in
the expected gradient; the latter describes the probability
of observing nonzero centered advantages.
Accordingly, SERA uses $\kappa$ as the allocation criterion
and $U$ as an observability safeguard.

\paragraph{Low-success budget concentration.}
For a fixed positive target exposure $c$, the unclipped
allocation in Equation~\ref{eq:closed-form-allocation}
satisfies
\begin{equation}
    N_q-1
    =\frac{c}{-\log(1-p_q)}
    \sim\frac{c}{p_q}
    \qquad (p_q\to0).
    \label{eq:app-small-p-demand}
\end{equation}
Maintaining a given fidelity therefore requires increasingly
many rollouts as the success probability approaches zero.
Under a fixed total budget, the common water level adjusts
to the full set of probabilities, so very low-success prompts
can absorb a large budget share and lower the common fidelity
attainable across the batch.
This motivates combining per-prompt count bounds with
observability-guided selection when distributing the budget.

\subsection{Comparison with Inverse-Fidelity Reweighting}
\label{app:inverse-fidelity}

For oracle $p$ and a fixed integer $N\ge2$,
$\kappa(p,N)>0$, so the corrected estimator
$\widetilde g:=\widehat g^{(N)}/\kappa(p,N)$ satisfies
$\mathbb E[\widetilde g]=g$.
Whenever the covariance exists,
\begin{equation}
    \operatorname{Cov}(\widetilde g)
    =\frac{1}{\kappa(p,N)^2}
       \operatorname{Cov}(\widehat g^{(N)}).
    \label{eq:app-inverse-fidelity-covariance}
\end{equation}
This correction rescales the sampled estimator.
It leaves the probability of observing a mixed group
unchanged, and all-correct or all-incorrect groups still
produce zero centered advantages.
For an individual prompt, increasing its rollout count
instead increases both $\kappa(p,N)$ and $U(p,N)$.
Rollout allocation therefore acts on the sampling process,
affecting both finite-rollout scaling and mixed-group
observability.

\section{Implementation Details}
\label{app:implementation-details}

\subsection{Historical Success-Probability Estimates}
\label{app:p-estimation}

We maintain historical success statistics using persistent
prompt identifiers.
For each prompt $q$, let $S_q^{(e)}$ and $F_q^{(e)}$ denote
the discounted success and failure counts available at the
start of epoch $e$.
Both are initialized to zero.
Let $n_q^{(e)}$ and $k_q^{(e)}$ be the numbers of responses
and successes collected during epoch $e$.
At the next epoch boundary, we update
\begin{equation}
    \begin{aligned}
        S_q^{(e+1)}
        &=\lambda S_q^{(e)}+k_q^{(e)},\\
        F_q^{(e+1)}
        &=\lambda F_q^{(e)}
          +n_q^{(e)}-k_q^{(e)},\\
        \widehat p_q^{(e+1)}
        &=\frac{\alpha_0+S_q^{(e+1)}}
          {\alpha_0+\beta_0+
           S_q^{(e+1)}+F_q^{(e+1)}}.
    \end{aligned}
    \label{eq:discounted-beta-update}
\end{equation}
Here $\alpha_0,\beta_0>0$ are fixed prior parameters and
$\lambda\in[0,1]$ controls the discount applied to historical
evidence.
The prior parameters themselves are not discounted.

The estimates $\widehat p_q^{(e)}$ remain fixed throughout
epoch $e$.
New outcomes are accumulated in pending counts and
incorporated at the next epoch boundary.
Thus, allocation decisions for the current responses depend
only on previously incorporated observations.
For a prompt receiving no responses during an epoch,
$n_q^{(e)}=k_q^{(e)}=0$, and only the historical evidence
is discounted.

\paragraph{Availability and uniform initialization.}
A prompt has an available estimate once observations for its
identifier have been incorporated at an epoch boundary.
The prior alone does not mark an unseen prompt as available.
If any prompt in a candidate batch lacks an available estimate,
we retain the full batch and assign $N_q=N_0$ to every prompt.
These uniform batches also contribute observations to the
historical estimator.
Previously observed prompts remain available even when they
receive no responses in a later epoch.

This procedure supplies smoothed historical estimates for
allocation without requiring additional preliminary rollouts
from the current batch.
Task-specific prior and discount parameters are reported in
Appendix~\ref{app:experimental-details}.

\subsection{Bounded Allocation and Exact-Budget Integerization}
\label{app:integerization}

For a candidate batch $\mathcal B$ of size $B$, the total
rollout budget is $M=BN_0$.
Given a retained set $\mathcal A$, we allocate this full budget
using the frozen success estimates $\widehat p_q$.
The integer count bounds satisfy
$2\le N_{\min}\le N_0\le N_{\max}$.
A retained set is feasible when
\begin{equation}
    |\mathcal A|N_{\min}
    \le M
    \le |\mathcal A|N_{\max}.
    \label{eq:app-allocation-capacity}
\end{equation}

\paragraph{Bounded continuous counts.}
Applying the allocation derived in
Appendix~\ref{app:waterline}, define
$h_q:=-\log(1-\widehat p_q)$ and solve
\begin{equation}
    x_q
    :=
    \operatorname{clip}\!\left(
        1+\frac{c_{\mathcal A}}{h_q},
        N_{\min},N_{\max}
    \right),
    \qquad
    \sum_{q\in\mathcal A}x_q=M.
    \label{eq:bounded-waterline-appendix}
\end{equation}
The sum of the clipped counts is continuous and nondecreasing
in $c_{\mathcal A}$.
For an interior budget, the implementation uses 100 bisection
iterations in log water level.
It returns the lower-budget side of the final bracket before
integerization.
At either endpoint of
Equation~\ref{eq:app-allocation-capacity}, the budget forces
every retained count to the corresponding bound.

The water level is recomputed whenever the retained set changes.
It is also solved with the bounds already included, so the
bounded allocation preserves the original budget.

\paragraph{Completing the integer budget.}
Let $\gamma_\star:=1-\exp(-c_{\mathcal A})$ denote the
interior fidelity target.
We first floor the continuous counts:
\begin{equation}
    n_q:=\lfloor x_q\rfloor,
    \qquad
    r:=M-\sum_{q\in\mathcal A}n_q.
    \label{eq:app-integer-residual}
\end{equation}
For every prompt with $n_q<N_{\max}$, compute the change
in squared fidelity error caused by one additional rollout:
\begin{equation}
    \Delta_q
    :=
    [\kappa(\widehat p_q,n_q+1)-\gamma_\star]^2
    -
    [\kappa(\widehat p_q,n_q)-\gamma_\star]^2.
    \label{eq:integer-marginal-error}
\end{equation}
Let $\mathcal R$ contain the $r$ below-cap prompts with the
smallest $\Delta_q$, breaking ties by their original
candidate-batch order.
The final counts are
\begin{equation}
    N_q=n_q+\mathbf1_{\{q\in\mathcal R\}},
    \qquad q\in\mathcal A.
    \label{eq:app-integer-completion}
\end{equation}
The marginal errors are computed once, and each selected
prompt receives one increment.
Every below-cap prompt is eligible, including a prompt whose
continuous count was already an integer.

This completion satisfies
\begin{equation}
    \sum_{q\in\mathcal A}N_q=M,
    \qquad
    N_{\min}\le N_q\le N_{\max}.
    \label{eq:app-integer-budget-guarantee}
\end{equation}
For any feasible one-increment completion
$z_q\in\{0,1\}$ with $\sum_q z_q=r$, the change in total
squared fidelity error is $\sum_q z_q\Delta_q$.
Selecting the smallest marginals therefore minimizes this
error within the class of one-increment completions.
We denote the complete bounded allocation and integerization
procedure by \textsc{Allocate}.

\subsection{Active-Prefix Search and Capacity Fallback}
\label{app:active-set-details}

For each candidate prompt, compute its estimated mixed-group
probability at the maximum permitted count:
\begin{equation}
    s_q:=\widehat U_q(N_{\max}),
    \qquad
    \widehat U_q(N)
    :=1-(1-\widehat p_q)^N-\widehat p_q^N.
    \label{eq:app-activation-score}
\end{equation}
We stably rank the prompts by decreasing $s_q$ and let
$\mathcal A_K$ denote the first $K$ prompts.
The minimum retained-set size required to absorb the budget
and the number passing the maximum-count screen are
\begin{equation}
    K_{\min}
    :=\left\lceil\frac{M}{N_{\max}}\right\rceil,
    \qquad
    K_{\mathrm{elig}}
    :=\bigl|\{q\in\mathcal B:s_q\ge u_0\}\bigr|.
    \label{eq:app-active-prefix-bounds}
\end{equation}
Since $M=BN_0$ and $N_{\min}\le N_0\le N_{\max}$,
every prefix size between $K_{\min}$ and $B$ satisfies
the count-capacity conditions.

For each tested prefix, we recompute its allocation using
the original budget $M$ and evaluate
\begin{equation}
    \min_{q\in\mathcal A_K}
        \widehat U_q(N_q^{(K)})
    \ge u_0-10^{-12}.
    \label{eq:active-prefix-predicate}
\end{equation}
This check uses the final integer counts returned by
\textsc{Allocate}.
The tolerance accommodates floating-point comparisons
near the threshold.

The implementation uses binary search over prefix sizes in
$[K_{\min},K_{\mathrm{elig}}]$ and returns the largest passing
prefix it encounters. Because integerization can break exact
monotonicity, this need not be the largest passing prefix over
all possible sizes.
If the maximum-count screen leaves fewer than $K_{\min}$
prompts, or no tested prefix passes the final-count check,
we allocate the budget to the top $K_{\min}$ prompts.
This capacity fallback preserves the budget and count bounds;
the returned prompts may fall below the observability threshold.
Inactive prompts receive zero rollouts.

\begin{algorithm}[t]
\caption{Observability-guided prefix search}
\label{alg:app-active-search}
\begin{algorithmic}[1]
\Require Candidate batch $\mathcal B$ with available estimates
         $\widehat{\mathbf p}$
\Require Budget $M=BN_0$, bounds $N_{\min},N_{\max}$,
         threshold $u_0$
\State Stably rank $\mathcal B$ by decreasing
       $\widehat U_q(N_{\max})$
\State Compute $K_{\min}$ and $K_{\mathrm{elig}}$ using
       Equation~\ref{eq:app-active-prefix-bounds}
\State $l\gets K_{\min}$;
       $h\gets K_{\mathrm{elig}}$;
       $K_{\mathrm{best}}\gets0$
\While{$l\le h$}
    \State $K\gets\lfloor(l+h)/2\rfloor$
    \State $\mathcal A_K\gets$ the top $K$ prompts
    \State $\mathbf N^{(K)}\gets
        \Call{Allocate}{
            \mathcal A_K,\widehat{\mathbf p},
            M,N_{\min},N_{\max}}$
    \If{$\min_{q\in\mathcal A_K}
          \widehat U_q(N_q^{(K)})\ge u_0-10^{-12}$}
        \State $K_{\mathrm{best}}\gets K$;
               $\mathbf N^{\mathrm{best}}\gets\mathbf N^{(K)}$
        \State $l\gets K+1$
    \Else
        \State $h\gets K-1$
    \EndIf
\EndWhile
\If{$K_{\mathrm{best}}>0$}
    \State $\mathcal A\gets\mathcal A_{K_{\mathrm{best}}}$;
           $\mathbf N\gets\mathbf N^{\mathrm{best}}$
\Else
    \State $\mathcal A\gets\mathcal A_{K_{\min}}$
    \State $\mathbf N\gets
        \Call{Allocate}{
            \mathcal A,\widehat{\mathbf p},
            M,N_{\min},N_{\max}}$
\EndIf
\State Set $N_q=0$ for $q\in\mathcal B\setminus\mathcal A$
\State \Return $\mathcal A,\mathbf N$
\end{algorithmic}
\end{algorithm}

\subsection{Loss Aggregation and Multiplicity Correction}
\label{app:loss-normalization}
\label{app:numerical-stabilization}

\paragraph{Correction for heterogeneous rollout counts.}
Each response from a retained prompt receives the weight
$w_q=N_0/N_q$.
Using the unstabilized centered advantage $a_{qi}$ from
Equation~\ref{eq:practical-maxrl}, response-mean score
aggregation gives
\begin{equation}
    \frac1M
    \sum_{q\in\mathcal A}\sum_{i=1}^{N_q}
        w_q a_{qi}s_{qi}
    =
    \frac1B
    \sum_{q\in\mathcal A}\widehat g_q^{(N_q)}.
    \label{eq:multiplicity-correction-identity}
\end{equation}
The equality follows from
$M=BN_0$ and $N_qw_q=N_0$.
Thus, heterogeneous rollout counts preserve the same total
response weight for each retained prompt.
Weights are evaluated only for prompts with $N_q>0$.

In the sequence-task implementation, the response weight
multiplies the centered advantage before the policy loss is
computed.
Reward means are computed within the original generation
groups before the responses are partitioned into PPO
minibatches and microbatches.
The resulting advantages and response weights remain attached
to their response records throughout subsequent batching.

\paragraph{Sequence and token normalization.}
For an aggregation unit containing responses
$i\in\mathcal I$, let $q_i$ be the associated prompt,
$m_{it}$ the valid-response-token mask, and $\ell_{it}$
the token policy-loss term before response weighting.
We use the following reductions:
\begin{align}
    L_{\mathrm{seq}}
    &=
    \frac{
        \sum_{i\in\mathcal I}\sum_t
        w_{q_i}m_{it}\ell_{it}
    }{|\mathcal I|L_{\mathrm{cap}}},
    \label{eq:seqnorm-loss}\\
    L_{\mathrm{token}}
    &=
    \frac{
        \sum_{i\in\mathcal I}\sum_t
        w_{q_i}m_{it}\ell_{it}
    }{
        \sum_{i\in\mathcal I}\sum_t m_{it}
    }.
    \label{eq:token-mean-loss}
\end{align}
Here $L_{\mathrm{cap}}$ is a fixed length normalizer
shared by all responses.
Both reductions retain the response weights in the numerator.
The token-mean denominator is the number of valid response
tokens in the aggregation unit.

\section{Extended Experimental Details}
\label{app:experimental-details}

\subsection{Experimental Setup}
\label{app:experimental-setup}

For all tasks, $B$ denotes the candidate-prompt batch size and
$N_0$ denotes the average number of generations per prompt,
giving a fixed rollout budget $M=BN_0$ per full batch.
Fixed MaxRL assigns $N_q=N_0$ to every prompt.
For \method{}, we use an observability threshold $u_0=0.05$
and active-prompt bounds $N_{\min}=2$ and $N_{\max}=4N_0$.
Inactive prompts receive no rollouts.
Each response from an active prompt is weighted by $N_0/N_q$.

Maze, SmolLM2, and Qwen use historical success-probability
estimates with a Beta$(0.5,0.5)$ prior and an evidence discount
of $0.75$ per epoch.
Estimates are updated at epoch boundaries and remain frozen
within each epoch.
Batches containing prompts without available historical
estimates use uniform allocation, $N_q=N_0$.
ImageNet instead uses detached exact correct-class
probabilities at each training step.

We follow the task-specific setup of MaxRL~\citep{tajwar2026maxrl},
with the main training and validation settings summarized in
Figures~\ref{fig:config-imagenet}--\ref{fig:prompt-qwen3}.
Generations per prompt denotes the average budget $N_0$;
the realized count $N_q$ can vary across prompts under \method{}.
Sequence lengths are measured in tokens.
Grad update per step denotes the number of optimizer updates
per RL training step.
Training Steps denotes the checkpoint used for comparison.
For sequence tasks, MaxRL and \method{} use the same token-mean
or Seqnorm loss aggregation within each controlled comparison.

\subsubsection{ImageNet}
\label{app:setup-imagenet}

ImageNet-1K is an image-classification task in which a model
assigns an input image to one of 1,000 categories.
We formulate classification as a one-step decision problem:
each image is a prompt, the policy samples a class label,
and the reward is one if the sampled label matches the
ground-truth class and zero otherwise.
This setting provides an exact success probability,
$p_q=\pi_\theta(y_q^\star\mid q)$, allowing us to study
rollout allocation without success-probability estimation error.

\begin{figure}[H]
    \centering
    \includegraphics[width=\linewidth]{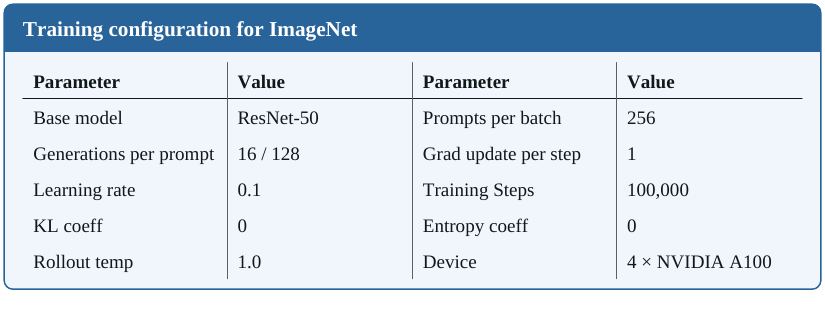}
    \caption{Training and validation settings for ImageNet.}
    \label{fig:config-imagenet}
\end{figure}

We train a randomly initialized ResNet-50 using 1,281,167
training images and 50,000 validation images.
Training inputs use random resized cropping to $224\times224$
and horizontal flipping; validation uses a $224\times224$
center crop.
We compare average rollout budgets $N_0\in\{16,128\}$ and
report the step-100,000 checkpoint.
Pass@$K$ is computed exactly as the validation-set mean of
$1-(1-p_q)^K$.
The main configuration is shown in
Figure~\ref{fig:config-imagenet}.

\subsubsection{Maze}
\label{app:setup-maze}

Maze is a sequential navigation task in which the policy must
generate a valid path from a designated starting position to
a goal while respecting the maze constraints.
A completed navigation sequence receives a binary reward
indicating whether it successfully solves the maze.
This task extends the one-step ImageNet setting to
multi-step generation with a verifiable outcome.

We use a fixed set of 7,424 training mazes and evaluate on
400 held-out mazes, all with grid size $17\times17$.
The initial policy is the step-1500 supervised fine-tuning
checkpoint of a Qwen2 decoder with approximately 3.94M
parameters.

\begin{figure}[H]
    \centering
    \includegraphics[width=\linewidth]{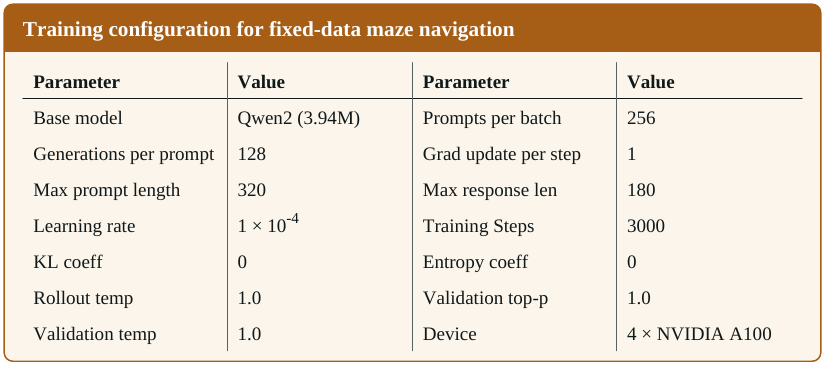}
    \caption{Training and validation settings for Maze.}
    \label{fig:config-maze}
\end{figure}

Both loss-normalization comparisons use the step-3000
checkpoint.
Evaluation samples 2,048 responses per maze.
The reported curves use empirical Pass@1 and
with-replacement bootstrap for $K\ge2$.
The main configuration is shown in
Figure~\ref{fig:config-maze}.

\begin{figure}[H]
    \centering
    \includegraphics[width=\linewidth]{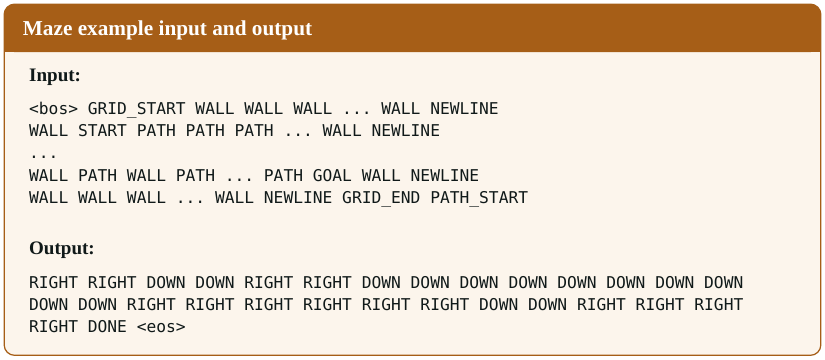}
    \caption{Example input and output for a $17\times17$ Maze.}
    \label{fig:prompt-maze}
\end{figure}

We represent each maze as a row-major sequence of symbolic cell
tokens and generate navigation actions after \texttt{PATH\_START}.
Figure~\ref{fig:prompt-maze} shows an example input and its
corresponding valid output, with part of the input omitted
for readability.

\subsubsection{SmolLM2 on GSM8K}
\label{app:setup-smollm}

GSM8K is a mathematical reasoning task consisting of
grade-school word problems that require multi-step arithmetic.
Given a problem, the model generates a solution and a final
numerical answer.
The final answer is checked against the reference answer
to obtain a binary correctness reward.
We use this setting to evaluate rollout allocation for
natural-language reasoning with a compact language model.

We initialize from SmolLM2-360M-Instruct and train on the
GSM8K training split.
Final evaluation is configured on the 1,209-question
GSM8K-Platinum set with 256 responses per question,
using the step-2000 checkpoint.

\begin{figure}[H]
    \centering
    \includegraphics[width=\linewidth]{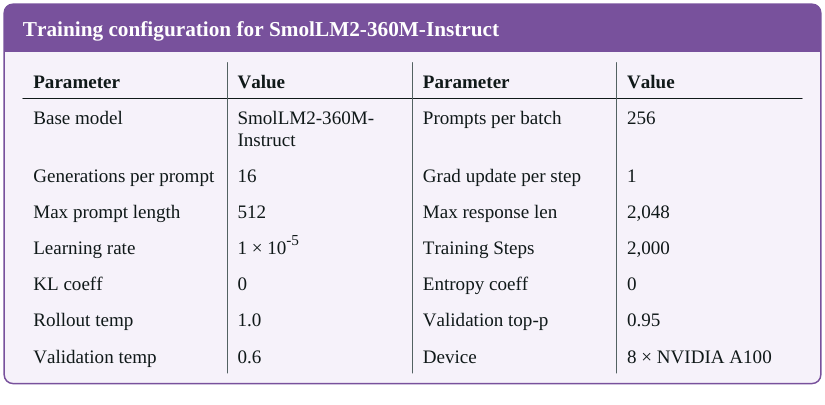}
    \caption{Training and validation settings for
    SmolLM2-360M-Instruct.}
    \label{fig:config-smollm}
\end{figure}

Pass@$K$ is computed using the combinatorial estimator.
All Pass@$K$ values for a checkpoint use the same response pool,
which is separate from the online validation pool.
The main configuration is shown in
Figure~\ref{fig:config-smollm}.

\begin{figure}[H]
    \centering
    \includegraphics[width=\linewidth]{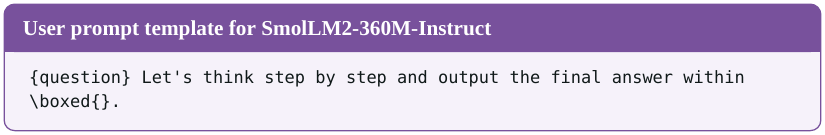}
    \caption{Prompt template for SmolLM2-360M-Instruct.}
    \label{fig:prompt-smollm}
\end{figure}

For both GSM8K training and GSM8K-Platinum evaluation, we append
the same reasoning and answer-format instruction to each question
and apply the SmolLM2 tokenizer's chat template.
Figure~\ref{fig:prompt-smollm} shows the prompt.

\subsubsection{Qwen}
\label{app:setup-qwen}

\begin{figure}[H]
    \centering
    \includegraphics[width=\linewidth]
        {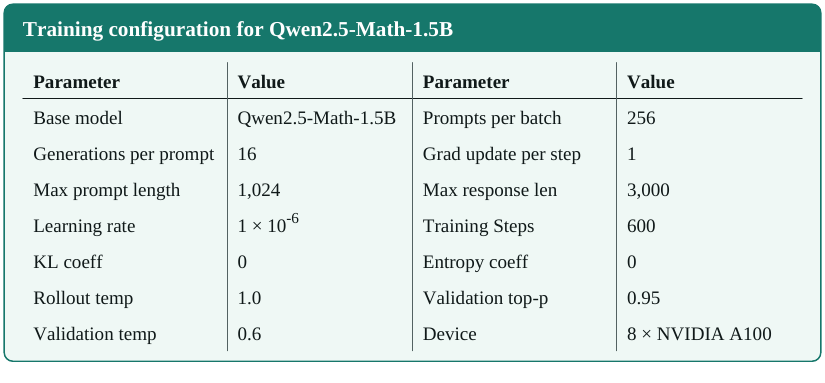}
    \caption{Training and validation settings for
    Qwen2.5-Math-1.5B.}
    \label{fig:config-qwen}
\end{figure}

\begin{figure}[H]
    \centering
    \includegraphics[width=\linewidth]
        {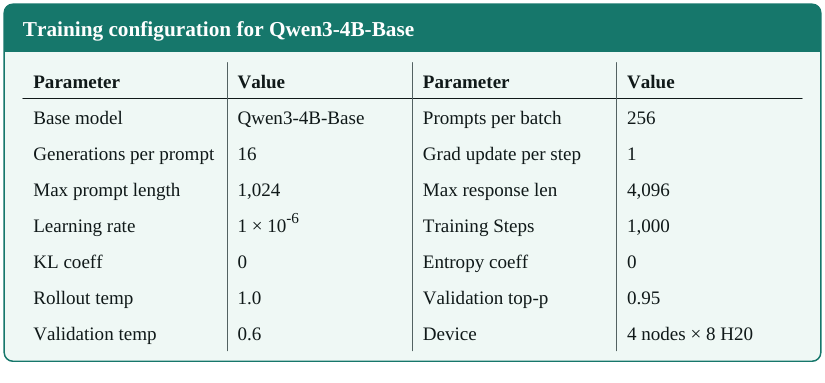}
    \caption{Training and validation settings for
    Qwen3-4B-Base.}
    \label{fig:config-qwen3-4b}
\end{figure}

The Qwen experiments evaluate mathematical reasoning
with Qwen2.5-Math-1.5B~\citep{yang2024qwen25math}
and Qwen3-4B-Base~\citep{yang2025qwen3}.
Qwen2.5-Math-1.5B is trained on MATH levels
3--5~\citep{hendrycks2021math}, while Qwen3-4B-Base
is trained on POLARIS-53K~\citep{an2025polaris}.
Both comparisons use token-mean loss aggregation.
For Qwen3-4B-Base, we follow the publicly released
MaxRL training configuration~\citep{tajwar2026maxrl},
retaining its data preprocessing, prompt template,
optimizer settings, generation limits, and total
training rollout budget.
Training and online validation settings are summarized
in Figures~\ref{fig:config-qwen}
and~\ref{fig:config-qwen3-4b}.

Final evaluation uses checkpoints at step $600$
for Qwen2.5-Math-1.5B and step $1{,}000$ for Qwen3-4B-Base.
For Qwen3-4B-Base, the GRPO and MaxRL baselines use
the publicly released checkpoints.
Qwen2.5-Math-1.5B is evaluated on BeyondAIME,
AIME 2025, MATH-500, and the English, text-only
mathematics subset of OlympiadBench.
Qwen3-4B-Base is evaluated on BeyondAIME,
AIME 2025, MATH-500, and Minerva Math. For both models, we sample $n=512$ responses per question
with temperature $0.6$ and top-$p$ $0.95$.
The maximum response length is $3{,}000$ tokens for
Qwen2.5-Math-1.5B and $4{,}096$ tokens for Qwen3-4B-Base.

We compute Pass@$K$ using the unbiased combination estimator,
\[
    \widehat{\mathrm{Pass@}K}
    =
    \frac{1}{Q}
    \sum_{q=1}^{Q}
    \left[
        1 -
        \frac{\binom{n-c_q}{K}}{\binom{n}{K}}
    \right],
\]
where $Q$ is the number of evaluation questions and
$c_q$ counts responses that pass the answer verifier
and have a generated length strictly below the
model-specific response limit.

\begin{figure}[H]
    \centering
    \includegraphics[width=\linewidth]{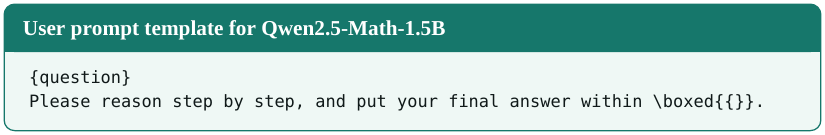}
    \caption{Prompt template for Qwen2.5-Math-1.5B.}
    \label{fig:prompt-qwen25}
\end{figure}

\begin{figure}[H]
    \centering
    \includegraphics[width=\linewidth]{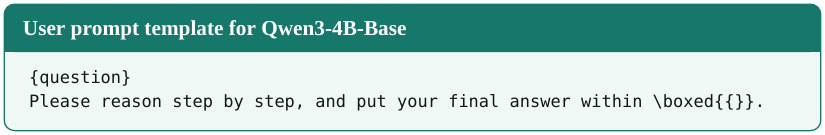}
    \caption{Prompt template for Qwen3-4B-Base.}
    \label{fig:prompt-qwen3}
\end{figure}

Figures~\ref{fig:prompt-qwen25} and~\ref{fig:prompt-qwen3} show the
resulting prompts for Qwen2.5-Math-1.5B and Qwen3-4B-Base, respectively.

\subsection{More Experimental Results}
\label{app:more-results}

\subsubsection{ImageNet}
\label{app:results-imagenet}

\paragraph{Final performance.}
Table~\ref{tab:app-imagenet-final} reports the final
Pass@$K$ profiles at step $100{,}000$, corresponding
to Figure~\ref{fig:imagenet-oracle-allocation} (c).
At $N_0=128$, \method{} outperforms MaxRL at every
reported $K$, increasing Pass@1 from $36.59\%$ to
$54.80\%$ and Pass@128 from $53.22\%$ to $87.51\%$.
These improvements substantially narrow the gap
to direct cross-entropy (CE) training across the
evaluated sampling budgets.
At $N_0=16$, \method{} also improves Pass@1 and
Pass@128, with gains extending through $K=512$.

\begin{table}[htbp]
    \centering
    \caption{\textbf{Final ImageNet Pass@$K$ (\%).}
    Results at step $100{,}000$, corresponding to
    Figure~\ref{fig:imagenet-oracle-allocation} (c).
    Bold indicates the higher value within each
    matched-budget MaxRL--\method{} pair.
    Direct CE provides the exact-likelihood
    training reference.}
    \label{tab:app-imagenet-final}
    \small
    \setlength{\tabcolsep}{2.3pt}
    \begin{tabular}{@{}llrrrrrrrrrrr@{}}
        \toprule
        $N_0$ & Method
        & $1$ & $2$ & $4$ & $8$ & $16$ & $32$
        & $64$ & $128$ & $256$ & $512$ & $1024$ \\
        \midrule
\textcolor{black!60}{--}
& \textcolor{black!60}{Direct CE}
& \textcolor{black!60}{60.81}
& \textcolor{black!60}{70.26}
& \textcolor{black!60}{77.38}
& \textcolor{black!60}{82.78}
& \textcolor{black!60}{86.95}
& \textcolor{black!60}{90.18}
& \textcolor{black!60}{92.68}
& \textcolor{black!60}{94.59}
& \textcolor{black!60}{96.05}
& \textcolor{black!60}{97.16}
& \textcolor{black!60}{98.00} \\
        \midrule
        $16$ & MaxRL
        & 4.43 & 4.97 & 5.37 & 5.71 & 6.05
        & 6.47 & 7.13 & 8.24 & 10.21 & 13.67
        & \textbf{19.35} \\
        $16$ & \method{}
        & \textbf{12.37} & \textbf{13.61}
        & \textbf{14.54} & \textbf{15.26}
        & \textbf{15.82} & \textbf{16.25}
        & \textbf{16.59} & \textbf{16.86}
        & \textbf{17.09} & \textbf{17.29}
        & 17.49 \\
        \midrule
        $128$ & MaxRL
        & 36.59 & 41.30 & 44.79 & 47.43 & 49.45
        & 51.01 & 52.24 & 53.22 & 54.00 & 54.62
        & 55.14 \\
        $128$ & \method{}
        & \textbf{54.80} & \textbf{63.53}
        & \textbf{70.26} & \textbf{75.46}
        & \textbf{79.52} & \textbf{82.75}
        & \textbf{85.37} & \textbf{87.51}
        & \textbf{89.25} & \textbf{90.65}
        & \textbf{91.79} \\
        \bottomrule
    \end{tabular}
\end{table}

\paragraph{Exact-gradient reference and measurement.}
ImageNet provides an exact reference for measuring
how closely sampled updates align with likelihood
optimization.
Let $\mathcal B$ denote the full candidate batch,
with $B=|\mathcal B|$, and let
$p_q(\theta)=\pi_\theta(y_q^\star\mid q)$
be the probability assigned to the ground-truth
class $y_q^\star$.
We define
\begin{equation}
    \begin{aligned}
        L_{\mathrm{CE}}(\theta;\mathcal B)
        &=-\frac{1}{B}
        \sum_{q\in\mathcal B}\log p_q(\theta), \\
        H_{\mathrm{CE}}
        &=\nabla_\theta L_{\mathrm{CE}}(\theta;\mathcal B).
    \end{aligned}
    \label{eq:app-imagenet-ce-reference}
\end{equation}
This reference covers the entire candidate batch,
including prompts assigned zero rollouts.
For the realized sampled training loss
$L_{\mathrm{sample}}$, incorporating the run's
allocation and loss normalization, let
$H_{\mathrm{sample}}=\nabla_\theta L_{\mathrm{sample}}$.
We measure gradient alignment as
\begin{equation}
    \operatorname{Cosine}
    =
    \frac{
        \langle H_{\mathrm{sample}},H_{\mathrm{CE}}\rangle
    }{
        \|H_{\mathrm{sample}}\|_2
        \|H_{\mathrm{CE}}\|_2
    }.
    \label{eq:app-imagenet-gradient-cosine}
\end{equation}

Both vectors use the loss-gradient convention and
cover all trainable model parameters.
Within each run, they are computed from the same
forward pass at the same current parameters and
inputs, after backpropagation and before gradient
clipping or optimizer operations, including
weight decay.
The diagnostic therefore measures sampled-to-exact
alignment along each method's own training trajectory.
The separately CE-trained model supplies the
performance reference in
Figure~\ref{fig:imagenet-oracle-allocation} (a--c),
while the CE reference gradient in
Figure~\ref{fig:imagenet-oracle-allocation} (d)
is computed at the corresponding RL model.

At $N_0=128$, diagnostics are recorded every
$1{,}000$ steps from step $1{,}000$ through
step $100{,}000$.
Across these $100$ measurements, \method{}
achieves a mean cosine of $0.6266$, compared
with $0.4317$ for MaxRL.
At the final checkpoint, the corresponding
values are $0.7960$ and $0.4447$.
These results directly demonstrate improved
alignment with the full-batch likelihood gradient
along the observed training trajectories.
Together with the validation results, they show
that \method{} improves both gradient alignment
and held-out performance under the same rollout
budget.

\paragraph{Accuracy of success-probability estimates.}
We use historical estimates to preserve flexibility in allocating
the current batch's rollout budget.
Under a fixed budget $M=BN_0$, a pilot stage assigns $m$ samples
to each prompt before adaptive allocation, committing $Bm$ rollouts
and leaving $B(N_0-m)$ for adaptive continuation.
For $N_0=16$ and $m=8$, half of the budget is therefore assigned
uniformly in advance.
These pilot samples remain within $M$ and are reused for training
on active prompts.
Historical estimation reuses past rollout outcomes without
requiring a current-batch pilot stage.

We assess estimation quality on ImageNet, where
$p_q=\pi_\theta(y_q^\star\mid q)$ is the exact probability of
the correct class.
The historical estimator uses the discounted Beta update in
Appendix~\ref{app:p-estimation}, with $\alpha_0=\beta_0=0.5$ and
$\lambda=0.75$, and remains frozen within each epoch using only
outcomes from earlier epochs.
For each batch, we compute the mean absolute error (MAE) and
Pearson's correlation coefficient $r$ between the estimated and
exact success probabilities across all candidate prompts.
We report the arithmetic mean of each metric over all batches
in the final epoch.
The historical estimator's MAE of $0.0386$ and mean Pearson
correlation of $0.8134$ indicate that it captures variation in
prompt-level success probabilities without current-batch pilot
sampling (Table~\ref{tab:app-pq-estimation}).

\begin{table}[htbp]
    \centering
    \caption{\textbf{Success-probability estimation on ImageNet.}
    Batch-level metrics averaged over the $5{,}005$ batches in
    epoch $20$ ($N_0=16$, nominal $B=256$, seed $69$).
    $m$ denotes current-batch pilot samples per prompt; pilot
    estimates use $\operatorname{clip}(K/8,1/16,15/16)$,
    where $K$ counts pilot successes.
    The runs have different allocation configurations and
    training trajectories.}
    \label{tab:app-pq-estimation}
    \setlength{\tabcolsep}{8pt}
    \begin{tabular}{@{}lrrr@{}}
        \toprule
        Estimator & $m$ & MAE $\downarrow$ & Pearson $r$ $\uparrow$ \\
        \midrule
        Pilot estimate      & $8$ & 0.0634 & 0.9879 \\
        Historical estimate & $0$ & 0.0386 & 0.8134 \\
        \bottomrule
    \end{tabular}
\end{table}

\subsubsection{Maze}
\label{app:results-maze}

\paragraph{Final performance.}
Table~\ref{tab:app-maze-final} reports the final
Pass@$K$ profiles at step $3{,}000$, complementing
Figure~\ref{fig:exp-maze-dynamics}.
Evaluation uses $400$ held-out mazes with
$2{,}048$ responses per maze.
We report both token-mean normalization and
fixed-length sequence normalization (Seqnorm),
together with an additional GRPO baseline
under token-mean normalization.
The base-model reference uses the step-0
evaluation of the shared SFT checkpoint.

Under token-mean normalization, \method{}
outperforms MaxRL at every reported $K\ge4$,
with the advantage increasing at larger
evaluation budgets.
Under Seqnorm, \method{} improves the entire
reported Pass@$K$ profile.
The Pass@128 gains over MaxRL are $4.84$ and
$3.28$ percentage points under token-mean and
Seqnorm, respectively.
These results demonstrate stronger solution
coverage under both normalizations, with Seqnorm
also improving single-sample performance.

\begin{table}[htbp]
    \centering
    \caption{\textbf{Final Maze Pass@$K$ (\%).}
    RL results are evaluated at step $3{,}000$,
    using $2{,}048$ responses per held-out maze.
    The gray row reports the pre-RL base-model
    reference at step $0$.
    Bold indicates the higher value within each
    MaxRL--\method{} pair using the same normalization.}
    \label{tab:app-maze-final}
    \begin{tabular}{@{}llrrrrrr@{}}
        \toprule
        Normalization & Method
        & $1$ & $2$ & $4$ & $8$ & $16$ & $32$ \\
        \midrule
        \textcolor{black!60}{--}
        & \textcolor{black!60}{Base model}
        & \textcolor{black!60}{1.27}
        & \textcolor{black!60}{2.49}
        & \textcolor{black!60}{4.75}
        & \textcolor{black!60}{8.83}
        & \textcolor{black!60}{15.50}
        & \textcolor{black!60}{25.17} \\
        \midrule
        Token-mean & GRPO
        & 34.00 & 34.80 & 35.43 & 36.01 & 36.61 & 37.28 \\
        Token-mean & MaxRL
        & \textbf{52.95} & \textbf{59.16}
        & 63.16 & 66.16 & 68.62 & 70.75 \\
        Token-mean & \method{}
        & 49.98 & 58.19
        & \textbf{63.85} & \textbf{68.06}
        & \textbf{71.49} & \textbf{74.45} \\
        \midrule
        Seqnorm & MaxRL
        & 52.80 & 59.42 & 63.74 & 66.96 & 69.63 & 71.94 \\
        Seqnorm & \method{}
        & \textbf{55.61} & \textbf{61.66}
        & \textbf{65.93} & \textbf{69.45}
        & \textbf{72.43} & \textbf{74.97} \\
        \bottomrule
    \end{tabular}

    \par\medskip

    \begin{tabular}{@{}llrrrrrr@{}}
        \toprule
        Normalization & Method
        & $64$ & $128$ & $256$ & $512$ & $1024$ & $2048$ \\
        \midrule
        \textcolor{black!60}{--}
        & \textcolor{black!60}{Base model}
        & \textcolor{black!60}{37.18}
        & \textcolor{black!60}{49.67}
        & \textcolor{black!60}{60.51}
        & \textcolor{black!60}{68.79}
        & \textcolor{black!60}{74.59}
        & \textcolor{black!60}{78.28} \\
        \midrule
        Token-mean & GRPO
        & 38.01 & 38.74 & 39.47 & 40.24 & 41.03 & 41.77 \\
        Token-mean & MaxRL
        & 72.60 & 74.31 & 75.82 & 76.98 & 77.79 & 78.38 \\
        Token-mean & \method{}
        & \textbf{76.98} & \textbf{79.15}
        & \textbf{81.06} & \textbf{82.58}
        & \textbf{83.75} & \textbf{84.48} \\
        \midrule
        Seqnorm & MaxRL
        & 73.83 & 75.51 & 76.93 & 78.04 & 78.86 & 79.62 \\
        Seqnorm & \method{}
        & \textbf{77.06} & \textbf{78.79}
        & \textbf{80.19} & \textbf{81.40}
        & \textbf{82.45} & \textbf{83.22} \\
        \bottomrule
    \end{tabular}
\end{table}

\subsubsection{SmolLM2 on GSM8K}
\label{app:results-smollm}

We extend the comparison in
Table~\ref{tab:exp-smollm-results} with validation
trajectories, complete Pass@$K$ profiles, and training
diagnostics for SmolLM2-360M-Instruct.
MaxRL and \method{} use matched training response
budgets within each loss normalization.

\begin{figure}[htpb]
    \centering
    \includegraphics[width=\linewidth]
        {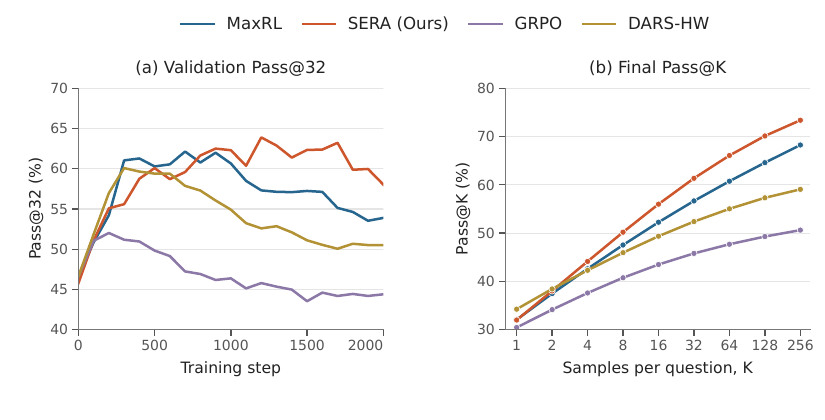}
    \caption{\textbf{SmolLM2 results on GSM8K-Platinum.}
    (a) Validation Pass@$32$ (bootstrap Best@$32$)
    during training.
    (b) Final Pass@$K$ at step $2{,}000$ from $256$
    responses per question.
    MaxRL, \method{} (Ours), and DARS-HW use Seqnorm;
    GRPO uses token-mean normalization.
    DARS-HW uses approximately $1.49\times$ as many
    training responses as the other methods.}
    \label{fig:app-smollm-coverage}
\end{figure}

Figure~\ref{fig:app-smollm-coverage} (a) tracks
validation Pass@$32$ for MaxRL, \method{}, GRPO,
and DARS-HW throughout training.
MaxRL, \method{}, and DARS-HW use fixed-length
sequence normalization (Seqnorm), while GRPO uses
token-mean normalization.
Coverage improves early in training for all four
methods.
During later training, MaxRL's coverage declines,
while \method{} maintains a higher Pass@$32$ and
finishes with stronger validation coverage.

Validation is performed every $100$ steps on
$1{,}209$ GSM8K-Platinum questions, with $32$
responses per question.
We plot the logged bootstrap Best@$32$ metric,
reported as Pass@$32$ by the trainer.
For each question, this statistic averages the
maximum binary reward over bootstrap samples of
$32$ responses drawn with replacement from its
validation pool, then averages across questions.

Figure~\ref{fig:app-smollm-coverage} (b) compares
the final Pass@$K$ profiles of MaxRL, \method{},
GRPO, and DARS-HW.
We evaluate the step-$2{,}000$ checkpoints using
a separate pool of $256$ responses per question
on the same evaluation set.
All values at
$K\in\{1,2,4,8,16,32,64,128,256\}$ are computed
from this pool using
\begin{equation}
    \widehat{\mathrm{Pass@}K}
    = \frac{1}{Q}\sum_{q=1}^{Q}
      \left(1-\frac{\binom{256-c_q}{K}}
                       {\binom{256}{K}}\right),
    \label{eq:app-smollm-passk}
\end{equation}
where $Q=1{,}209$, $c_q$ is the number of correct
responses for question $q$, and
$\binom{a}{K}=0$ for $K>a$.

\method{} achieves the highest Pass@$K$ among
the four methods at every displayed $K\ge4$.
Compared with MaxRL, it improves coverage at
every displayed $K\ge2$ while retaining nearly
identical Pass@1 under the same training response
budget.
It also outperforms GRPO across the entire
displayed range of evaluation budgets.

Table~\ref{tab:app-smollm-uncertainty} quantifies
uncertainty in the final Seqnorm comparison.
We use $10{,}000$ paired bootstrap resamples of
evaluation questions, keeping the checkpoints and
response pools fixed, and report pointwise $95\%$
percentile confidence intervals.
The intervals for \method{}'s gains over MaxRL
are above zero at all three reported budgets.

\begin{table}[htbp]
    \centering
    \caption{\textbf{SmolLM2 uncertainty analysis.}
    $\Delta$ denotes \method{} minus MaxRL at step
    $2{,}000$, in percentage points.
    CI denotes the question-paired $95\%$ confidence interval.}
    \label{tab:app-smollm-uncertainty}
    \small
    \setlength{\tabcolsep}{8pt}
    \begin{tabular}{lccc}
        \toprule
        & Pass@32 & Pass@128 & Pass@256 \\
        \midrule
        $\Delta$
        & $+4.69$ & $+5.54$ & $+5.13$ \\
        $95\%$ CI
        & $[2.32,\,7.04]$
        & $[3.09,\,8.01]$
        & $[2.48,\,7.86]$ \\
        \bottomrule
    \end{tabular}
\end{table}

\subsubsection{Mathematical reasoning with Qwen}
\label{app:results-qwen}

Table~\ref{tab:app-qwen-passk} reports the numerical
Pass@$K$ profiles of MaxRL and \method{} from
Figure~\ref{fig:exp-qwen}, together with the pretrained
base model as a reference.
All results use the evaluation protocol described in
Appendix~\ref{app:setup-qwen}.

\begin{table}[htbp]
\centering
\caption{\textbf{Qwen2.5-Math-1.5B Pass@$K$ (\%).}
MaxRL and \method{} are evaluated at step $600$ using $512$
sampled responses per question. Gray rows show the pretrained base model under the same protocol. Bold indicates the higher unrounded value between MaxRL and
\method{} for each benchmark and $K$.}
\label{tab:app-qwen-passk}
\small
\setlength{\tabcolsep}{2.5pt}
\renewcommand{\arraystretch}{1.1}
\begin{tabular}{@{}llrrrrrrrrrr@{}}
\toprule
Dataset & Method
& $1$ & $2$ & $4$ & $8$ & $16$
& $32$ & $64$ & $128$ & $256$ & $512$ \\
\midrule

\multirow{3}{*}{\shortstack[l]{BeyondAIME}}
    & \textcolor{black!60}{Base}
    & \textcolor{black!60}{0.73}
    & \textcolor{black!60}{1.41}
    & \textcolor{black!60}{2.61}
    & \textcolor{black!60}{4.57}
    & \textcolor{black!60}{7.36}
    & \textcolor{black!60}{10.96}
    & \textcolor{black!60}{15.49}
    & \textcolor{black!60}{21.20}
    & \textcolor{black!60}{28.13}
    & \textcolor{black!60}{35.00} \\
    & MaxRL
    & \textbf{3.29} & \textbf{5.25} & \textbf{7.71}
    & 10.82 & \textbf{14.82}
    & 19.53 & 24.78 & 30.53 & 36.53 & 42.00 \\
    & \method{}
    & 3.07 & 5.06 & 7.64
    & \textbf{10.82} & 14.76
    & \textbf{19.55} & \textbf{25.36} & \textbf{32.26}
    & \textbf{40.14} & \textbf{49.00} \\
\midrule

\multirow{3}{*}{\shortstack[l]{AIME 2025}}
    & \textcolor{black!60}{Base}
    & \textcolor{black!60}{3.26}
    & \textcolor{black!60}{6.01}
    & \textcolor{black!60}{10.36}
    & \textcolor{black!60}{16.17}
    & \textcolor{black!60}{22.59}
    & \textcolor{black!60}{29.21}
    & \textcolor{black!60}{36.42}
    & \textcolor{black!60}{44.21}
    & \textcolor{black!60}{53.12}
    & \textcolor{black!60}{63.33} \\
    & MaxRL
    & 10.05 & 15.64 & 21.65 & 27.37 & 33.02
    & 38.86 & 44.49 & 49.91 & 56.26 & 63.33 \\
    & \method{}
    & \textbf{10.18} & \textbf{16.28} & \textbf{23.04}
    & \textbf{29.19} & \textbf{34.77}
    & \textbf{40.46} & \textbf{46.30} & \textbf{52.85}
    & \textbf{61.26} & \textbf{70.00} \\
\midrule

\multirow{3}{*}{\shortstack[l]{MATH-500}}
    & \textcolor{black!60}{Base}
    & \textcolor{black!60}{26.70}
    & \textcolor{black!60}{40.03}
    & \textcolor{black!60}{54.22}
    & \textcolor{black!60}{67.28}
    & \textcolor{black!60}{77.93}
    & \textcolor{black!60}{85.68}
    & \textcolor{black!60}{90.80}
    & \textcolor{black!60}{94.10}
    & \textcolor{black!60}{96.25}
    & \textcolor{black!60}{97.60} \\
    & MaxRL
    & \textbf{73.91} & 80.55 & 85.07 & 88.12 & 90.43
    & 92.36 & 93.94 & 95.34 & 96.66 & 97.80 \\
    & \method{}
    & 73.89 & \textbf{80.68} & \textbf{85.34}
    & \textbf{88.55} & \textbf{91.04}
    & \textbf{93.16} & \textbf{94.85} & \textbf{96.20}
    & \textbf{97.29} & \textbf{98.00} \\
\midrule

\multirow{3}{*}{\shortstack[l]{OlympiadBench}}
    & \textcolor{black!60}{Base}
    & \textcolor{black!60}{20.35}
    & \textcolor{black!60}{29.70}
    & \textcolor{black!60}{39.17}
    & \textcolor{black!60}{47.75}
    & \textcolor{black!60}{55.28}
    & \textcolor{black!60}{61.64}
    & \textcolor{black!60}{66.68}
    & \textcolor{black!60}{70.83}
    & \textcolor{black!60}{74.77}
    & \textcolor{black!60}{78.67} \\
    & MaxRL
    & 36.05 & 43.32 & 49.84 & 55.74 & 61.05
    & 65.67 & 69.65 & 73.27 & 76.68 & 79.85 \\
    & \method{}
    & \textbf{36.20} & \textbf{43.54} & \textbf{50.19}
    & \textbf{56.37} & \textbf{62.09}
    & \textbf{67.12} & \textbf{71.35} & \textbf{74.91}
    & \textbf{78.01} & \textbf{80.89} \\
\bottomrule
\end{tabular}
\end{table}

Using the same question-paired bootstrap procedure,
Table~\ref{tab:app-qwen-uncertainty} reports uncertainty
in the final Qwen2.5-Math-1.5B comparisons.
The Pass@32 intervals are above zero on AIME 2025,
MATH-500, and OlympiadBench, while intervals become
wider at larger evaluation budgets.

\begin{table}[htbp]
    \centering
    \caption{\textbf{Qwen2.5-Math-1.5B uncertainty.}
    Each cell reports the difference between SERA and
MaxRL at step 600 (top), with its pointwise $95\%$ paired-bootstrap confidence interval (bottom), in
percentage points. Differences are computed from unrounded scores before rounding.}
    \label{tab:app-qwen-uncertainty}
    \small
    \setlength{\tabcolsep}{5pt}
    \renewcommand{\arraystretch}{1.2}
    \begin{tabular}{lcccc}
        \toprule
        Dataset & Pass@32 & Pass@128 & Pass@256 & Pass@512 \\
        \midrule
        BeyondAIME
        & \makecell{$+0.01$\\$[-1.59,\,1.56]$}
        & \makecell{$+1.72$\\$[-1.50,\,4.98]$}
        & \makecell{$+3.60$\\$[-1.21,\,8.50]$}
        & \makecell{$+7.00$\\$[0.00,\,15.00]$} \\
        \addlinespace[2pt]
        AIME 2025
        & \makecell{$+1.61$\\$[0.12,\,3.11]$}
        & \makecell{$+2.94$\\$[-0.82,\,7.27]$}
        & \makecell{$+5.00$\\$[-0.21,\,11.67]$}
        & \makecell{$+6.67$\\$[0.00,\,16.67]$} \\
        \addlinespace[2pt]
        MATH-500
        & \makecell{$+0.80$\\$[0.23,\,1.43]$}
        & \makecell{$+0.86$\\$[0.02,\,1.77]$}
        & \makecell{$+0.63$\\$[-0.26,\,1.58]$}
        & \makecell{$+0.20$\\$[-0.80,\,1.20]$} \\
        \addlinespace[2pt]
        OlympiadBench
        & \makecell{$+1.45$\\$[0.86,\,2.07]$}
        & \makecell{$+1.64$\\$[0.81,\,2.51]$}
        & \makecell{$+1.33$\\$[0.19,\,2.50]$}
        & \makecell{$+1.04$\\$[-0.59,\,2.81]$} \\
        \bottomrule
    \end{tabular}
\end{table}

Table~\ref{tab:app-qwen3-passk} reports the full Pass@$K$
results for the checkpoints at the step $1{,}000$.
Each question uses $512$ sampled responses.
At $K=512$, \method{} improves over MaxRL by $8.00$
percentage points on BeyondAIME and $4.78$ points on Minerva Math.

\begin{table}[htbp]
\centering
\caption{\textbf{Qwen3-4B-Base Pass@$K$ (\%).}
Gray rows show Base; bold marks the higher unrounded score
between MaxRL and \method{}.}
\label{tab:app-qwen3-passk}
\small
\setlength{\tabcolsep}{2.5pt}
\renewcommand{\arraystretch}{1.1}
\begin{tabular}{@{}llrrrrrrrrrr@{}}
\toprule
Dataset & Method
& $1$ & $2$ & $4$ & $8$ & $16$
& $32$ & $64$ & $128$ & $256$ & $512$ \\
\midrule

\multirow{3}{*}{\shortstack[l]{BeyondAIME}}
    & \textcolor{black!60}{Base}
    & \textcolor{black!60}{4.02}
    & \textcolor{black!60}{6.16}
    & \textcolor{black!60}{8.51}
    & \textcolor{black!60}{11.08}
    & \textcolor{black!60}{14.33}
    & \textcolor{black!60}{18.61}
    & \textcolor{black!60}{23.90}
    & \textcolor{black!60}{30.35}
    & \textcolor{black!60}{37.75}
    & \textcolor{black!60}{45.00} \\
    & MaxRL
    & 6.97 & 10.53 & 14.46 & 18.62 & 23.28
    & 28.50 & 34.08 & 40.27 & 47.50 & 55.00 \\
    & \method{}
    & \textbf{7.78} & \textbf{11.34} & \textbf{15.78}
    & \textbf{21.08} & \textbf{27.17}
    & \textbf{34.01} & \textbf{41.39} & \textbf{48.81}
    & \textbf{55.91} & \textbf{63.00} \\
\midrule

\multirow{3}{*}{\shortstack[l]{AIME 2025}}
    & \textcolor{black!60}{Base}
    & \textcolor{black!60}{6.82}
    & \textcolor{black!60}{11.78}
    & \textcolor{black!60}{18.33}
    & \textcolor{black!60}{25.23}
    & \textcolor{black!60}{31.48}
    & \textcolor{black!60}{37.23}
    & \textcolor{black!60}{43.08}
    & \textcolor{black!60}{49.90}
    & \textcolor{black!60}{57.41}
    & \textcolor{black!60}{63.33} \\
    & MaxRL
    & 16.37 & 22.75 & 28.50 & 33.84 & 39.43
    & 45.34 & 51.72 & 59.17 & 66.77 & \textbf{73.33} \\
    & \method{}
    & \textbf{18.37} & \textbf{23.85} & \textbf{29.30}
    & \textbf{35.38} & \textbf{41.88}
    & \textbf{48.40} & \textbf{54.97} & \textbf{61.20}
    & \textbf{67.61} & \textbf{73.33} \\
\midrule

\multirow{3}{*}{\shortstack[l]{MATH-500}}
    & \textcolor{black!60}{Base}
    & \textcolor{black!60}{67.40}
    & \textcolor{black!60}{78.81}
    & \textcolor{black!60}{85.37}
    & \textcolor{black!60}{89.43}
    & \textcolor{black!60}{92.21}
    & \textcolor{black!60}{94.21}
    & \textcolor{black!60}{95.66}
    & \textcolor{black!60}{96.68}
    & \textcolor{black!60}{97.41}
    & \textcolor{black!60}{98.00} \\
    & MaxRL
    & 81.52 & \textbf{87.83} & 91.20 & 93.13 & 94.48
    & 95.61 & 96.61 & 97.44
    & \textbf{98.00} & \textbf{98.40} \\
    & \method{}
    & \textbf{81.71} & 87.60 & \textbf{91.32}
    & \textbf{93.72} & \textbf{95.29}
    & \textbf{96.33} & \textbf{97.00} & \textbf{97.45}
    & 97.79 & 98.00 \\
\midrule

\multirow{3}{*}{\shortstack[l]{Minerva Math}}
    & \textcolor{black!60}{Base}
    & \textcolor{black!60}{25.72}
    & \textcolor{black!60}{33.95}
    & \textcolor{black!60}{41.55}
    & \textcolor{black!60}{48.12}
    & \textcolor{black!60}{53.37}
    & \textcolor{black!60}{57.65}
    & \textcolor{black!60}{61.23}
    & \textcolor{black!60}{63.95}
    & \textcolor{black!60}{65.74}
    & \textcolor{black!60}{66.91} \\
    & MaxRL
    & 37.30 & 44.01 & 49.06 & 53.03 & 56.34
    & 59.11 & 61.39 & 63.34 & 64.93 & 66.18 \\
    & \method{}
    & \textbf{38.55} & \textbf{44.91} & \textbf{50.18}
    & \textbf{54.27} & \textbf{57.38}
    & \textbf{60.02} & \textbf{62.52} & \textbf{65.11}
    & \textbf{67.96} & \textbf{70.96} \\
\bottomrule
\end{tabular}
\end{table}

Table~\ref{tab:app-qwen3-uncertainty} quantifies uncertainty
in the differences between \method{} and MaxRL.
We use $10{,}000$ paired-bootstrap resamples of evaluation questions and report the $2.5$th and $97.5$th percentiles.
The same question indices are used for both methods
and all values of $K$, and differences are computed before rounding.

\begin{table}[htbp]
\centering
\caption{\textbf{Qwen3-4B-Base uncertainty.}
\method{} minus MaxRL in percentage points (top),
with pointwise $95\%$ paired-bootstrap intervals (bottom). Differences are computed from unrounded scores before rounding.}
\label{tab:app-qwen3-uncertainty}
\small
\setlength{\tabcolsep}{5pt}
\renewcommand{\arraystretch}{1.2}
\begin{tabular}{lcccc}
\toprule
Dataset & Pass@32 & Pass@128 & Pass@256 & Pass@512 \\
\midrule
BeyondAIME
    & \makecell{$+5.51$\\$[1.69,\,9.54]$}
    & \makecell{$+8.54$\\$[3.42,\,13.78]$}
    & \makecell{$+8.41$\\$[2.19,\,14.74]$}
    & \makecell{$+8.00$\\$[0.00,\,17.00]$} \\
\addlinespace[2pt]
AIME 2025
    & \makecell{$+3.06$\\$[-5.74,\,11.67]$}
    & \makecell{$+2.03$\\$[-9.04,\,13.86]$}
    & \makecell{$+0.84$\\$[-9.70,\,11.88]$}
    & \makecell{$+0.00$\\$[-10.00,\,10.00]$} \\
\addlinespace[2pt]
MATH-500
    & \makecell{$+0.72$\\$[-0.05,\,1.55]$}
    & \makecell{$+0.00$\\$[-0.67,\,0.60]$}
    & \makecell{$-0.21$\\$[-0.89,\,0.39]$}
    & \makecell{$-0.40$\\$[-1.20,\,0.40]$} \\
\addlinespace[2pt]
Minerva Math
    & \makecell{$+0.91$\\$[-1.57,\,3.47]$}
    & \makecell{$+1.77$\\$[-0.50,\,4.14]$}
    & \makecell{$+3.03$\\$[0.60,\,5.56]$}
    & \makecell{$+4.78$\\$[1.84,\,8.09]$} \\
\bottomrule
\end{tabular}
\end{table}

\subsubsection{Training Dynamics and Allocation Behavior}
\label{app:training-dynamics}

\begin{figure}[ht]
    \centering
    \includegraphics[width=\linewidth]
        {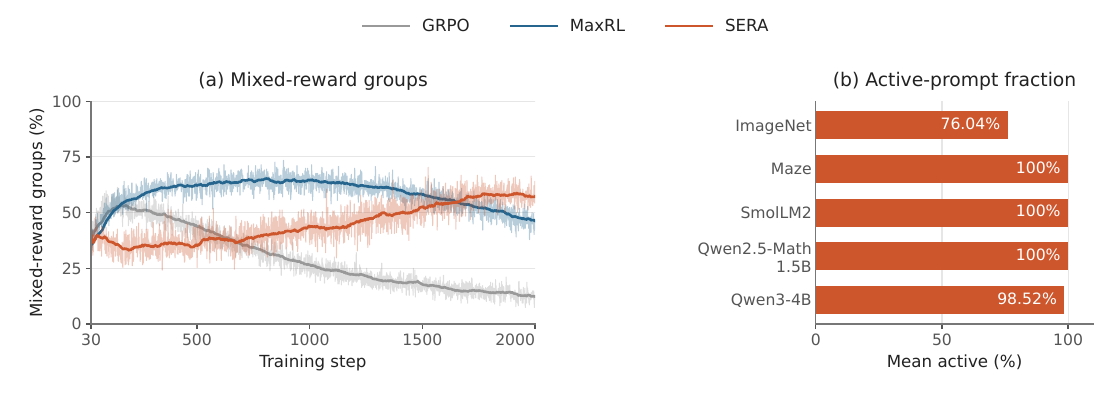}
    \caption{\textbf{Mixed-reward groups and active prompts.}
    (a) Mixed-reward group fraction on SmolLM2, using Seqnorm
    for MaxRL and \method{} and token-mean normalization
    for GRPO.
    Faint and solid lines show raw values and trailing
    $50$-step means.
    (b) Mean active-prompt fraction for \method{} across tasks.}
    \label{fig:training-mechanisms}
\end{figure}

In the analyzed SmolLM2-360M-Instruct run,
\method{} achieves stronger multi-sample coverage
than MaxRL with fewer cumulative mixed-reward groups
and all candidate prompts remaining active.
Figure~\ref{fig:training-mechanisms} (a) compares
both methods under matched rollout budgets and
fixed-length sequence normalization (Seqnorm),
with token-mean GRPO as a reference.
Mixed-reward groups contain both correct and
incorrect responses and yield nonzero centered advantages.
Their fraction under \method{} rises late in training
and overtakes MaxRL's, despite a lower cumulative
count over steps $30$--$2{,}000$.
Complementary evidence comes from ImageNet,
where \method{} improves alignment with the exact
cross-entropy gradient
(Figure~\ref{fig:imagenet-oracle-allocation} (d)),
supporting the motivation to account for finite-rollout
gradient scaling when allocating rollouts.

Figure~\ref{fig:training-mechanisms} (b) reports the
mean per-step active-prompt fraction $|\mathcal A_t|/B_t$,
where $\mathcal A_t$ is the active prompt set and
$B_t$ is the candidate-prompt batch size.
All candidate prompts remain active throughout the
analyzed Maze, SmolLM2, and Qwen2.5-Math-1.5B windows.
These task gains therefore occur with rollouts
redistributed across the full candidate batch.
The observability criterion $U_q$ guides prompt activation,
while rollout allocation targets equalization of the
gradient-scaling coefficient $\kappa$ within the active set.

\begin{figure}[t]
    \centering
    \includegraphics[width=\linewidth]
        {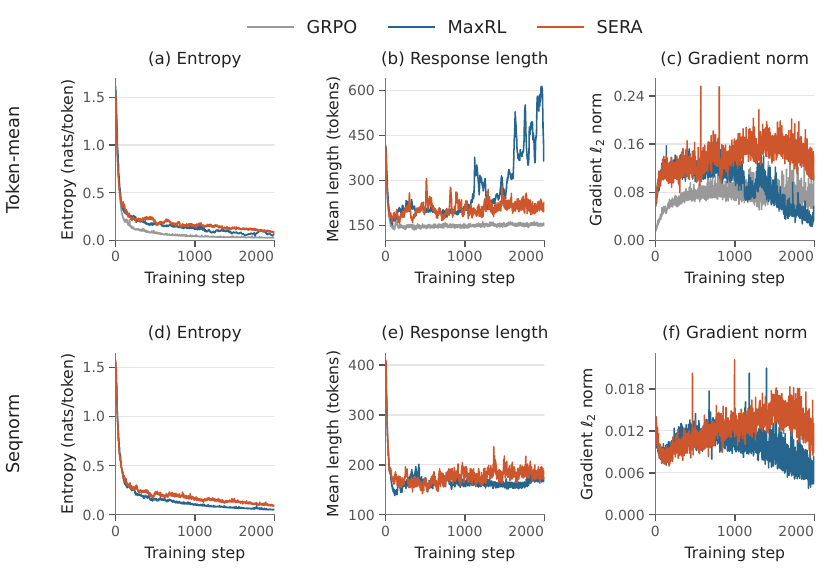}
    \caption{\textbf{SmolLM2 training diagnostics.}
    Rows show token-mean normalization (top) and
    Seqnorm (bottom); GRPO appears only in the top row.
    Columns show token entropy, mean response length,
    and the actor gradient $\ell_2$ norm.}
    \label{fig:app-smollm-training}
\end{figure}

Figure~\ref{fig:app-smollm-training} presents complementary
training diagnostics on SmolLM2-360M-Instruct.
MaxRL and \method{} use matched rollout budgets within
each normalization, with GRPO included in the token-mean
comparison.
\method{} maintains higher late-training token entropy
than MaxRL under both normalizations and than GRPO
under token-mean normalization.
These entropy trends accompany its stronger
multi-sample coverage.
Under token-mean normalization, MaxRL's mean response
length increases markedly late in training, while
GRPO and \method{} produce shorter responses.
Under Seqnorm, the response lengths of MaxRL and
\method{} remain closer throughout training.
Entropy is measured in nats per valid response token,
and response length is averaged over generated responses.
The actor gradient $\ell_2$ norm is measured after
gradient accumulation and before clipping.

\begin{figure}[t]
    \centering
    \includegraphics[width=\linewidth]
        {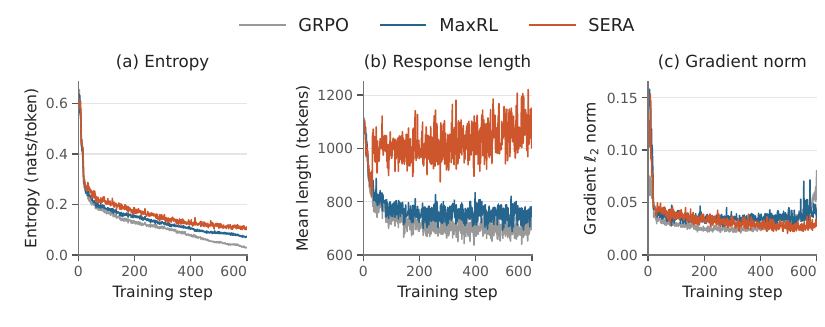}
    \caption{\textbf{Qwen2.5-Math-1.5B training diagnostics.}
    (a) Token entropy.
    (b) Mean response length.
    (c) Actor gradient $\ell_2$ norm.
    GRPO, MaxRL, and \method{} use token-mean normalization.}
    \label{fig:app-qwen25-training}
\end{figure}

Figure~\ref{fig:app-qwen25-training} extends these diagnostics
to Qwen2.5-Math-1.5B over the $600$-step training window.
All three methods use token-mean normalization and the
same total rollout budget per step.
\method{} maintains higher late-training token entropy
than GRPO and MaxRL, accompanied by longer generated responses.

\end{document}